\documentclass[letterpaper, 10 pt, journal, twoside]{support/IEEEtran}
\usepackage{times}
\usepackage[pdftex]{graphicx}
\usepackage{subfigure}
\usepackage{amsmath,amssymb,amsopn,amstext,amsfonts}
\usepackage{cancel}
\usepackage[space]{cite}
\usepackage{pdfsync}
\usepackage{balance}
\usepackage{color}
\usepackage{mathtools}
\usepackage{algpseudocode}
\usepackage[ruled,vlined,linesnumbered]{algorithm2e}
\algnewcommand\algorithmicforeach{\textbf{for each}}
\algdef{S}[FOR]{ForEach}[1]{\algorithmicforeach\ #1\ \algorithmicdo}

\usepackage{bm}
\usepackage{subfigure}
\usepackage{diagbox}
\usepackage{float}
\usepackage{epstopdf}
\usepackage{pifont}
\usepackage{amsmath}
\usepackage{multirow}
\usepackage{url}
\usepackage{verbatim}
\usepackage{footmisc}
\usepackage{mathabx}	
\usepackage{physics}
\usepackage[linkcolor=black,citecolor=black,urlcolor=black,colorlinks=true]{hyperref}
\usepackage{subfigure}
\usepackage{color}
\usepackage{amssymb,mathrsfs,amsmath}
\usepackage[skip=3pt, font={small}]{caption} 
\usepackage{tabularx}
\usepackage{tikz}
\usepackage{amsthm}
\usepackage{wrapfig}
\newcommand{\todo}[1]{\textcolor{red}{\emph{\bf#1}}}
\newcommand{\HL}[1]{{#1}}
\newcommand{\VIDEO}[1]{{\it#1}}

\usepackage{tabulary}
\usepackage{booktabs}
\usepackage{siunitx}
\usepackage{accents}
\usepackage{pdflscape}
\usepackage[flushleft]{threeparttable}
\usepackage{stmaryrd} 
\newcommand{\removelatexerror}{\let\@latex@error\@gobble}

\newtheoremstyle{mystyle}
{}
{}
{\itshape}
{}
{\bfseries}
{.}
{ }
{\thmname{#1}\thmnumber{ #2}\thmnote{ (#3)}}

\theoremstyle{mystyle}

\graphicspath{{figures/}}
\DeclareGraphicsExtensions{.png,.jpg,.eps,.pdf}

\def\BibTeX{{\rm B\kern-.05em{\sc i\kern-.025em b}\kern-.08em
		T\kern-.1667em\lower.7ex\hbox{E}\kern-.125emX}}
\title{\Large \textbf{ImMesh: An Immediate LiDAR Localization and Meshing Framework}
}
\author{
	Jiarong Lin$^*$, Chongjian Yuan$^*$, Yixi Cai, Haotian Li, Yunfan Ren, Yuying Zou, Xiaoping Hong, and Fu Zhang 
    \thanks{Manuscript received February 5, 2023; revised July 23, 2023; accepted September 18, 2023. This work is supported by the University Grants Committee of Hong Kong General Research Fund (project number 17206421) and DJI Donation. \textit{(Corresponding author: Fu Zhang.)}}
    \thanks{$^*$These two authors contribute equally to this work.}
	\thanks{J. Lin, C. Yuan, Y. Cai and F. Zhang are with the Department of Mechanical Engineering, The University of Hong Kong, Hong Kong SAR, China. {\tt\footnotesize $\{$jiarong.lin, ycj1, yixicai, haotianl, renyf, zyycici, fuzhang$\}$@connect.hku.hk}}
    
    \thanks{X. Hong are with the School of System Design and Intelligent Manufacturing, Southern University of Science and Technology, Shenzhen, People’s Republic of China. {\tt\footnotesize $\{$hongxp$\}$@sustech.edu.cn}}
}%

\makeatletter
\let\@oldmaketitle\@maketitle
\renewcommand{\@maketitle}{\@oldmaketitle
	\centering
	\setcounter{figure}{0}
	\begin{minipage}{1.0\linewidth}
		\includegraphics[width=1.0\textwidth]{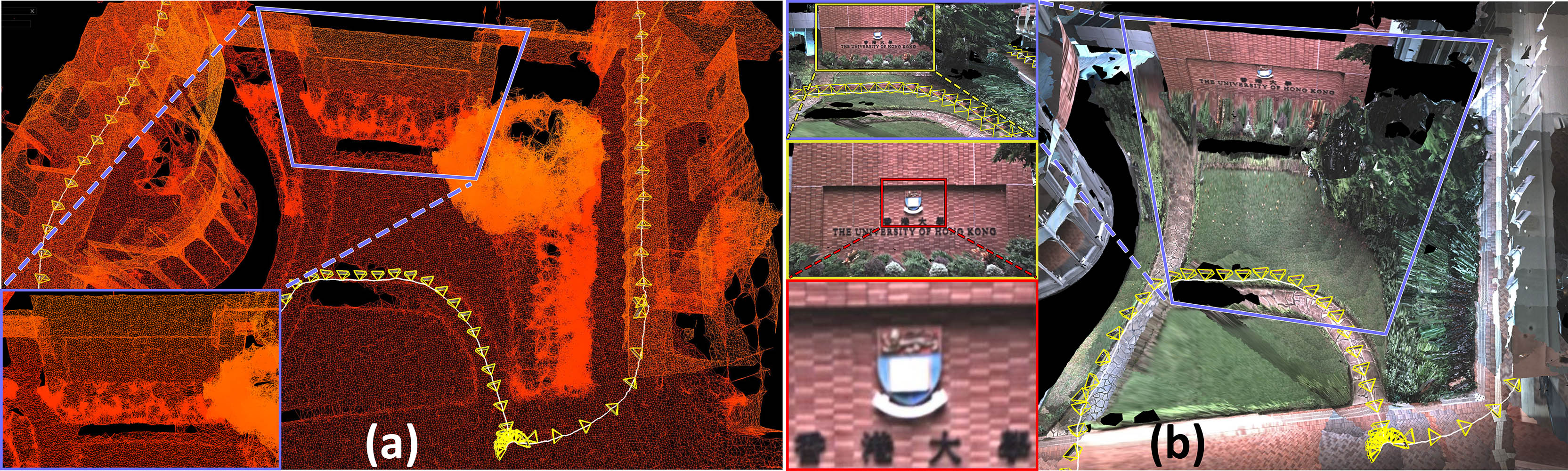}
		\includegraphics[width=1.0\textwidth]{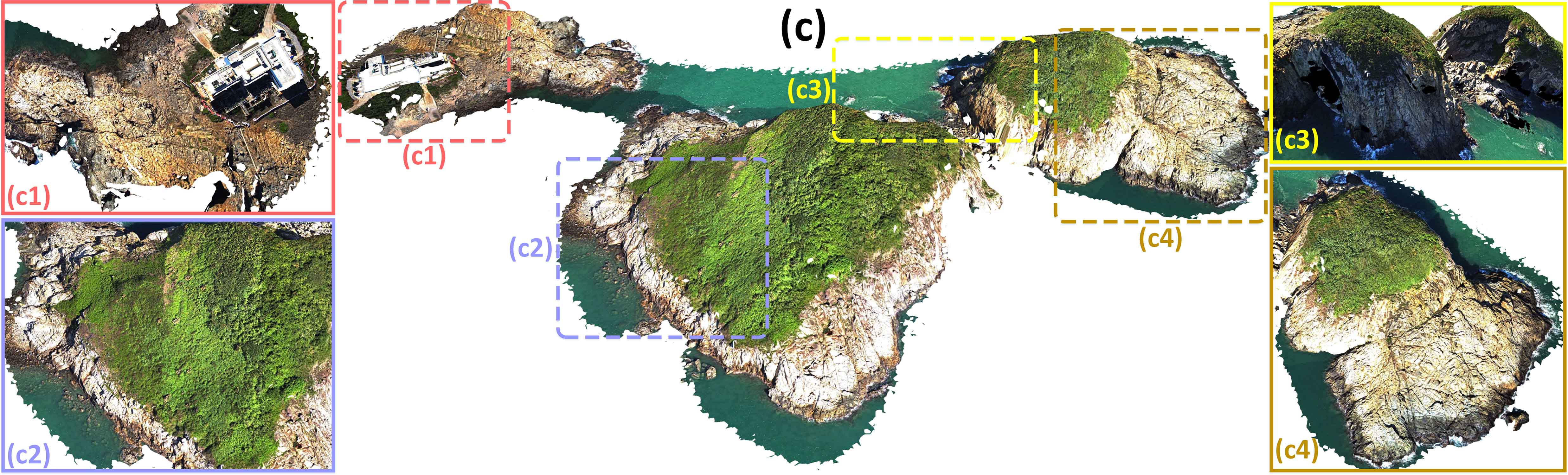}
        \vspace{-0.35cm}
        \captionof{figure}{(a) shows the triangle mesh that is online reconstructed by our proposed work ImMesh, where the white path is our sampling trajectory, and the yellow frustums are the estimated sensor pose. In (b), we use the estimated camera poses (the yellow frustums) of R$^3$LIVE for texturing the mesh with the collected images. Based on ImMesh, we developed a lossless texture reconstruction application, with one of our results shown in (c). Our accompanying video that shows details of this work is available on YouTube: \href{https://youtu.be/pzT2fMwz428}{\tt{youtu.be/pzT2fMwz428}}.}
        \label{fig_cover}
        \vspace{-0.1cm}
	\end{minipage}
	\vspace{-1cm}
}
	
\makeatother

\begin{document}
	\maketitle	
	\begin{abstract}
        In this paper, we propose a novel LiDAR(-inertial) odometry and mapping framework to achieve the goal of simultaneous localization and meshing in real-time. This proposed framework termed ImMesh comprises four tightly-coupled modules: receiver, localization, meshing, and broadcaster. The localization module first utilizes the preprocessed sensor data from the receiver, estimates the sensor pose online by registering LiDAR scans to maps, and dynamically grows the map. Then, our meshing module takes the registered LiDAR scan for incrementally reconstructing the triangle mesh on the fly. Finally, the real-time odometry, map, and mesh are published via our broadcaster. The primary contribution of this work is the meshing module, which represents a scene by an efficient voxel structure, performs fast finding of voxels observed by new scans, and incrementally reconstructs triangle facets in each voxel. This voxel-wise meshing operation is delicately designed for the purpose of efficiency; it first performs a dimension reduction by projecting 3D points to a 2D local plane contained in the voxel, and then executes the meshing operation with pull, commit and push steps for incremental reconstruction of triangle facets. To the best of our knowledge, this is the first work in literature that can reconstruct online the triangle mesh of large-scale scenes, just relying on a standard CPU without GPU acceleration. To share our findings and make contributions to the community, we make our code publicly available on our GitHub: \href{https://github.com/hku-mars/ImMesh}{\tt github.com/hku-mars/ImMesh}.
	\end{abstract}

\setlength{\textfloatsep}{-0.2cm}

\section{Introduction}


\IEEEPARstart{R}{ecently}, the wide emergence of 3D applications such as metaverse\cite{mystakidis2022metaverse, wang2022survey}, VR/AR\cite{cipresso2018past}, video games, and physical simulator\cite{shah2018airsim, song2021flightmare} has enriched human lifestyle and boosted productive efficiency by providing a virtual environment that \HL{resembles} the real world. These applications are built upon triangle meshes that represent the complex geometry of real-world scenes. Triangle mesh is the collection of vertices and triangle facets, which serves as a fundamental tool for object modeling in most existing 3D applications. It can not only significantly simplify the process and boost the speed of rendering\cite{laine2011high, akenine2019real} and ray-tracing\cite{arvo2013graphics}, but also play an irreplaceable role in collision detection\cite{jimenez20013d, ericson2004real}, rigid-body dynamics\cite{featherstone2014rigid, baraff1997introduction}, dense mapping and surveying \cite{schonberger2016structure}, sensor simulation \cite{kong2022marsim, wang2020tartanair}, etc. However, most of the existing mesh is manufactured by skillful 3D modelers with the help of computer-aided design (CAD) software (e.g., Solidworks\cite{solidworks2005solidworks}, blender\cite{blender2018blender}, etc.), which limits the mass production of large-scene meshing. Hence, developing an efficient mesh method that could reconstruct large scenes in real-time draws increasing research interests and serves as a hot topic in the community of 3D reconstruction. 

Performing mesh reconstruction in real-time is particularly important in practical usages. Firstly, online mesh reconstruction makes data collection effective by providing a live preview, which is essential to give users a reference. Especially for non-expert users, a live preview can provide feedback about which parts of the scene have already been reconstructed in good quality and where additional data is needed. Secondly, online mesh reconstruction can immediately output the mesh of the scene once data collection is complete, saving additional post-processing time of offline mesh reconstruction and boosting the productivity of mass production. Thirdly, it is particularly important for those real-time applications, especially fully autonomous robotic applications. A real-time update of mesh can provide better maps with denser representation and higher accuracy, enabling the agent to navigate itself better.

Reconstructing the mesh of large scenes from sensor measurements in real-time remains one of the most challenging problems in computer graphics, 3D vision, and robotics, which require reconstructing the surfaces of scenes with triangle facets adjacently connected by edges. This challenging problem needs to build the geometry structure with very high accuracy, and the triangle facet should be reconstructed on surfaces that actually exist in the real world. Besides, a good mesh reconstruction method should also suppress the appearance of holes on the reconstructed surface and avoid the reconstruction of triangle silver (i.e., the noodle-like triangles with an acute shard angle). Real-time mesh reconstruction in large scenes is even more challenging as it further requires the reconstruction to operate efficiently and incrementally. 

In this work, we propose a real-time mesh reconstruction framework termed ImMesh to achieve the goal of simultaneous localization and meshing on the fly. ImMesh is a well-engineered system comprised of four tightly-coupled modules delicately designed for efficiency and accuracy. Among them, we implement a novel mesh reconstruction method in our meshing module.  Specifically, our meshing module first utilizes the voxels for partitioning the 3D space and allows fast finding of voxels that contain points of new scans. Then, the voxel-wise 3D meshing problem is converted into a 2D one by performing dimension reduction for efficient meshing. Finally, the triangle facets are incrementally reconstructed with the voxel-wise mesh pull, commit and push steps. To the best of our knowledge, this is the first work in literature to reconstruct the triangle mesh of large-scale scenes online with a standard CPU. The main contributions of our work are:
\begin{itemize}
    \item We propose ImMesh, a novel SLAM framework designed to achieve simultaneous localization and mesh reconstruction using a LiDAR sensor. ImMesh is built upon our previous work VoxelMap \cite{voxelmap}, and incorporates a novel mesh reconstruction method. This proposed approach can efficiently and incrementally reconstruct the mesh of scenes online, achieving real-time performance in large-scale scenarios on a standard desktop CPU.
    \item We comprehensively evaluated ImMesh's runtime performance and meshing accuracy using real-world and synthetic data, by comparing our runtime performance and meshing accuracy against existing baselines to assess its effectiveness.
    \item We additionally demonstrate how real-time meshing can be applied in potential applications by presenting two practical examples: point cloud reinforcement and losless texture reconstruction (see Fig. \ref{fig_cover}(b and c)).
    \item We make ImMesh publicly available on our GitHub: \href{https://github.com/hku-mars/ImMesh}{\tt github.com/hku-mars/ImMesh} for sharing our findings and making contributions to the community, 
\end{itemize}

\section{Related Works}
In this section, we discuss the related works of mesh reconstruction based on 3D point clouds, which are closely related to this work.  Depending on whether the reconstruction processes can perform online, we categorize existing mesh reconstruction methods into two classes: offline methods and online methods.

\subsection{Offline mesh reconstruction}
The offline methods usually require a global map in prior, for example, the full registered point cloud of the scene. Then, a global mesh reconstruction process is used to build the mesh. In this category, the most notable works include: methods based on Poisson surface reconstruction (Poisson-based), and methods based on Delaunay tetrahedralization (i.e., 3D Delaunay triangulation) and graph cut (Delaunay-based).

\subsubsection{Poisson surface reconstruction (Poisson-based)}
Given a set of 3D points with oriented normals that are sampled on the surface of a 3D model, the basic idea of Poisson surface reconstruction \cite{kazhdan2006poisson, kazhdan2013screened} is to cast the problem of mesh reconstruction  as an optimization problem, which solves for an approximate indicator function of the inferred solid whose gradient best matches the input normals. Then, the continuous isosurface (i.e., the triangle mesh) is extracted from the indicator function using the method \cite{wilhelms1992octrees, shekhar1996octree}, similar to adaptations of the Marching Cubes \cite{lorensen1987marching} with octree representations. 

Benefiting from this implicit representation, where the mesh is extracted from the indicator function instead of being estimated directly, Poisson surface reconstruction can produce a watertight manifold mesh and is resilient to scanner noise, misalignment, and missing data. Hence, in the communities of graphics and vision, these types of methods \cite{kazhdan2006poisson, kazhdan2013screened, kazhdan2020poisson} have been widely used for reconstructing the mesh from given 3D scanned data. 

\subsubsection{Delaunay triangulation and graph cut (Delaunay-based)} 
In the category of offline mesh reconstruction methods, approaches \cite{labatut2007efficient, litvinov2013incremental, jancosek2014exploiting} based on Delaunay tetrahedralization and graph cut have also been widely used for generating the mesh, relying on the reconstructed 3D point cloud and the sensor's poses. The basic idea of this class of methods is first to build a tetrahedral decomposition of 3D space by computing the 3D Delaunay triangulation of the 3D point set. Then, the Delaunay tetrahedra were labeled as two classes (i.e., ``inside'' or ``outside") with the globally optimal label assignment (i.e., the graph cut). Finally, the triangle mesh can be extracted as the interface between these two classes.

Besides these two classes of methods, there are other offline mesh reconstruction methods, such as the ball-pivoting algorithm \cite{bernardini1999ball}. This algorithm works by pivoting a ball of fixed radius around each point in the point cloud and constructing a triangle whenever three balls overlap. \cite{cao2010point} involves extracting the curve skeleton using Laplacian-based contraction, and then reconstructing the surface with the skeleton-assisted topology. However, these methods are often not the first choice due to various limitations such as robustness, accuracy, and efficiency when compared to Poisson- and Delaunay-based methods \cite{wang2018lidar}.


Unlike these offline mesh reconstruction methods, our proposed work ImMesh can perform online in an incremental manner without the complete point cloud of the scene. Besides, ImMesh also achieves a satisfactory meshing accuracy that is higher than Poisson-based methods and slightly lower than Delaunay-based methods (see our experimental results in Section~\ref{section_exp_3}).

\subsection{Online mesh reconstruction}
    \subsubsection{Voxel volume-based methods (TSDF-based)}
    The online mesh reconstruction method is predominated by TSDF-based methods, which represent the scene in a voxel volumetric theme. {These methods implicitly reconstruct the mesh in a two-step pipeline, which first establishes the truncated signed distance to the closest surface of voxels, then extracts the continuous triangle mesh by leveraging the Marching Cubes algorithm \cite{lorensen1987marching} from volumes}. TSDF-based methods are popularized by KinectFusion \cite{newcombe2011kinectfusion}, with many follow-up works focused on scaling this approach to larger scenes \cite{chen2013scalable, niessner2013real}, adding multi-resolution capability \cite{ kahler2015hierarchical, Vespa_RAL2018}, and improving efficiency \cite{kahler2015very, klingensmith2015chisel, oleynikova2017voxblox}. Since these classes of methods can be easily implemented with parallelism, they can achieve real-time performance with the acceleration of GPUs.  


    Compared to these methods, our work ImMesh shows several advantages: Firstly, in ImMesh, the triangle mesh is directly reconstructed from the point cloud in one step, while for TSDF-based methods, the mesh is implicitly built in a two-step pipeline (i.e., SDF update followed by a mesh extraction). Secondly, ImMesh can output the mesh in scan rate (i.e., sensor sampling rate), while the mesh extraction of TSDF-based methods is usually at a lower rate. Thirdly, ImMesh achieves real-time performance by running on a standard CPU, while TSDF-based methods need GPU acceleration for real-time SDF updates. Lastly, TSDF-based methods require adequate observation for the calculation of the SDF of each voxel w.r.t. the \HL{closest} surface, which needs the data to be sampled by a depth sensor of high resolution and moving at a low speed. On the contrary, our work exploits high-accuracy LiDAR points for meshing and is robust to points data of low density.
    
    

    \subsubsection{Surfel-based mesh reconstruction}
    Besides TSDF-based methods, another popular approach is representing the scene with a set of points or surfels (e.g., oriented discs). For example, in work \cite{niessner2013real, lefloch2017comprehensive, lefloch2015anisotropic}, the maps are reconstructed with point-based representation, and its ``surface" is rendered with the approaches of ``point-based rendering" that originated from the communities of computer graphics\cite{weise2009hand, rusinkiewicz2002real, habbecke2007surface}. Besides, in work\cite{bodenmueller2009streaming}, the high-quality map is reconstructed with surfel-based representations (i.e., use patches). Such forms of mapping representation are popularized in works\cite{whelan2015elasticfusion, whelan2016elasticfusion, gao2019surfelwarp, schops2019surfelmeshing}.  To reconstruct a dense map, these classes of methods need a large number of points or tiny patches to represent the surface of the models, which is an inefficient representation with high usage of system memory and computation resources. In contrast, our work reconstructs the surface of models with triangle mesh, which uses triangle facets of proper size adjacently connected by edges. It is the most efficient solid-model representation that has been widely adopted in most modern 3D software.
   
Compared with the works reviewed above, our proposed work is in a class by itself, which contains the following advantages:
\begin{itemize} 
    \item It is an online mesh reconstruction method that reconstructs the triangle mesh in an incremental manner. It can achieve real-time performance in large-scale scenes (e.g., traveling length reaches $\SI{7.5}{\kilo\meter}$) by just running on a standard desktop CPU.
    \item It explicitly reconstructs the triangle mesh by directly taking the registered LiDAR points as meshing vertices, performing the voxel-wise meshing operation as each new LiDAR scan is registered.
    \item It is delicately designed for the purpose of efficiency and achieves satisfactory meshing precision comparable to existing high-accuracy offline methods.
\end{itemize}

\begin{figure*}
	\centering
        \vspace{-0.1cm}
	\includegraphics[width=1.00\linewidth]{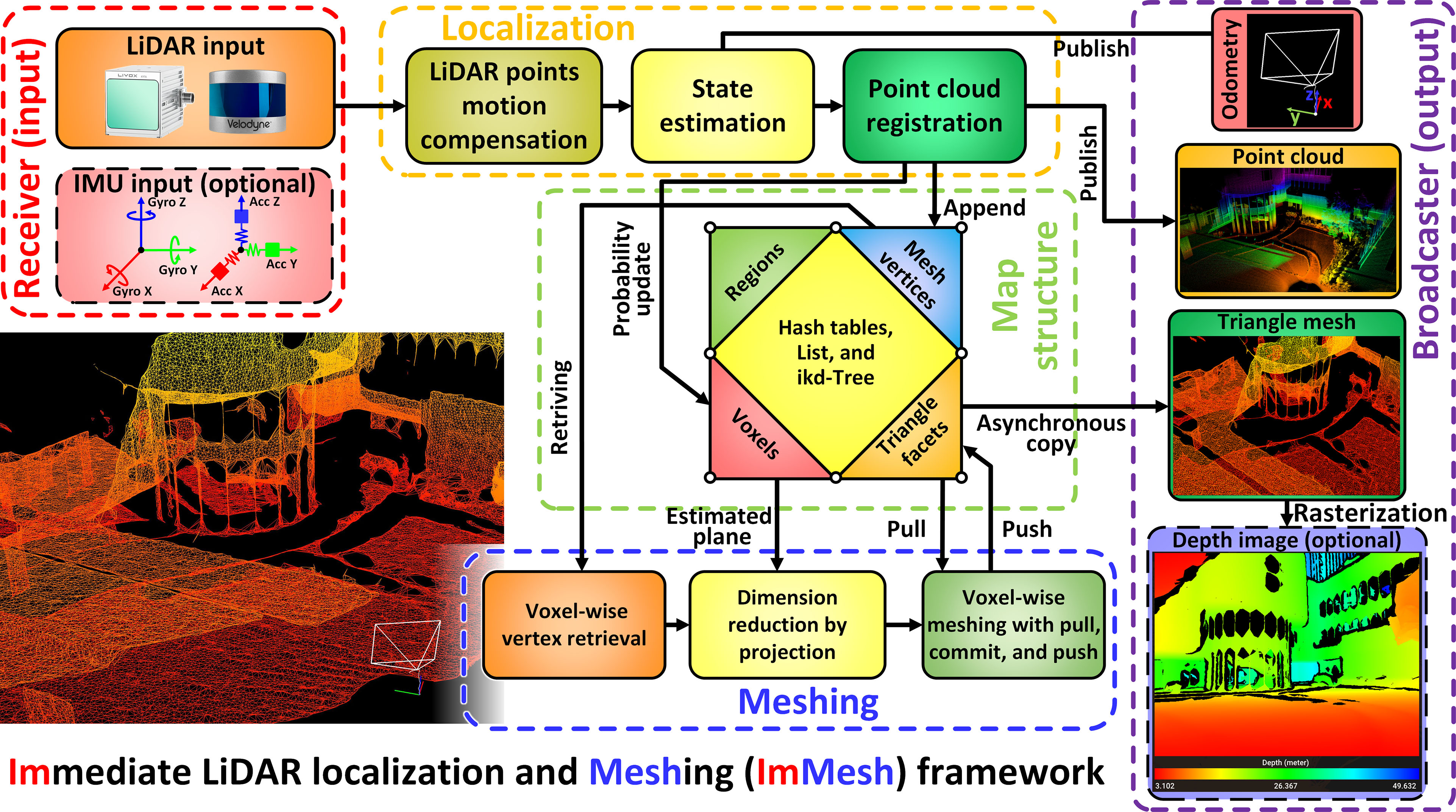}
	\caption{This figure shows the overview of our proposed work ImMesh, which utilizes the raw input sensor data to achieve the goal of simultaneous localization and meshing. It is constituted by four tightly-coupled modules and a map structure, from left (input) to right (output) are: \textit{receiver} (in red), \textit{localization} (in orange),  \textit{map structure} (in green), \textit{meshing} (in blue) and \textit{broadcaster} (in purple).}
	\label{fig_overview}
    \vspace{-0.5cm}
\end{figure*}

\section{System overview}
Fig.~\ref{fig_overview} depicts the overview of our proposed system (ImMesh), which consists of a map structure and four modules that work jointly to achieve the goal of simultaneous localization and meshing in real-time. As shown in Fig.~\ref{fig_overview}, from left to right are: \textit{receiver} (in red), \textit{localization} (in orange), \textit{map structure} (in green),  \textit{meshing} (in blue) and \textit{broadcaster} (in purple).

In the rest sections, we will first introduce our \textit{map structures} in Section \ref{sect_meth_mapping}, showing the detail of the data structures used in other modules. Next, we will introduce our receiver and localization module in Section \ref{sect_meth_receiver_locaization}. Then, we will present how our \textit{meshing} modules work in Section \ref{sect_meth_mesh}. Finally, in Section \ref{sect_meth_broadcaster}, we will introduce the \textit{broadcaster} module, which publishes the localization and meshing results to other applications.

\section{Map structure}\label{sect_meth_mapping}

As shown by the \textit{map structure} (in green) in Fig. \ref{fig_overview}, we designed four data types, including mesh vertices, triangle facets, regions, and voxels, as well as two data structures: a hash table for efficient data lookup and an incremental kd-tree (ikd-tree) for $k$ nearest neighbors (kNN) search and downsampling.

\begin{figure*}[t]
	\centering
	\includegraphics[width=1.00\linewidth]{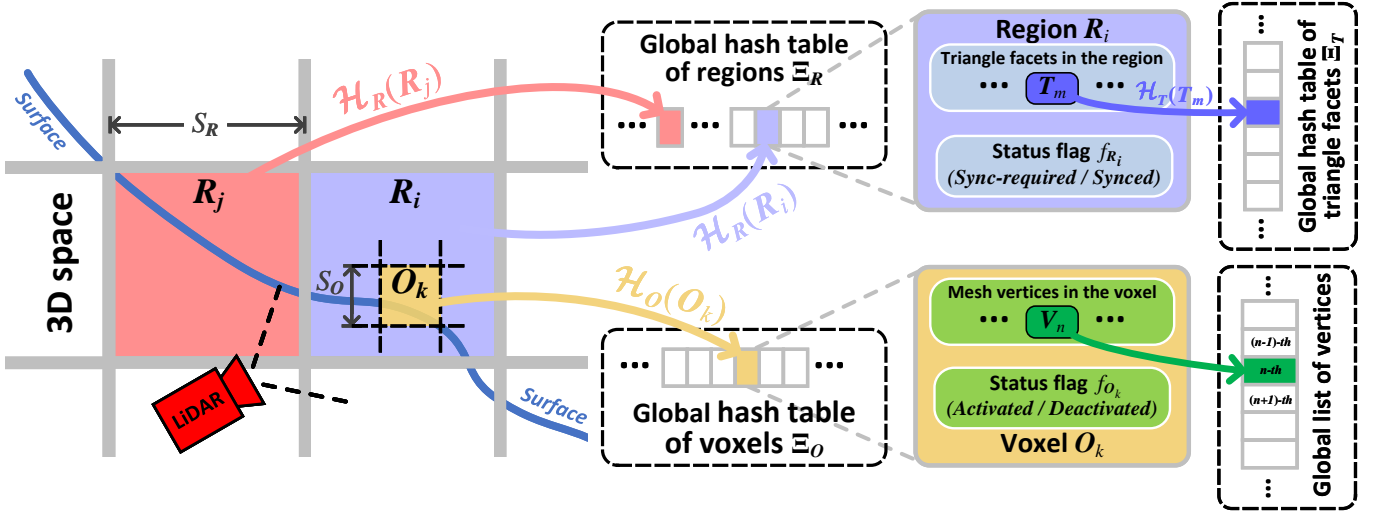}
	\caption{In ImMesh, we partition the 3D space into two types of volumetric grids: regions and voxels. Triangle facets are stored inside the regions, and mesh vertices are stored inside the voxels. Additionally, we maintain three hash tables to facilitate efficient lookup of these data types. }
	\label{fig_hierarchical_voxels}
    \vspace{-0.2cm}
\end{figure*}

The relationship among these map structures is depicted in Fig. \ref{fig_hierarchical_voxels}, where we partition the 3D space into two types of volumetric grids: regions and voxels. Triangle facets are stored inside the regions containing them and are also indexed in a global hash table of triangle facets, and mesh vertices are stored inside the voxels containing them and are also indexed in a global list of vertices. Additionally, we maintain two hash tables to facilitate the efficient lookup of regions and voxels.

\subsection{Data types: Region, voxel, triangle facet, and mesh vertex}\label{sect_hierachical_voxel}
        
\subsubsection{Region $\mathbf{R}$} Region have a much larger size $S_{\mathbf{R}}$ (e.g., $S_{\mathbf{R}} = \SI{10.0}{\meter}$) compared to voxel's size $S_{\mathbf{O}}$ (e.g., $S_{\mathbf{O}} = \SI{0.4}{\meter}$). They contain triangle facets whose centers are located inside, allowing for the \textit{broadcaster} to asynchronously copy these triangle facets. Additionally, each region has a status flag $f_{\mathbf{R}}$ to identify its syncing status, which can be either \textit{Sync-required} or \textit{Synced}. This status indicates the update flag related to the data synchronization of triangle facets.
    
\subsubsection{Voxel $\mathbf{O}$} Voxels enable the \textit{meshing} module to efficiently retrieve all in-voxel mesh vertices for voxel-wise meshing operations. Each voxel $\mathbf{O}_i$ also has a status flag $f_{\mathbf{O}}$ indicating whether it has new points appended. Specifically, $\mathbf{O}_i$ is marked as \textit{Activated} if new mesh vertices are registered from the latest LiDAR scan. The \textit{Activated} flag is reset to \textit{Deactivated} after the voxel-wise meshing operation has been performed on this voxel.

\subsubsection{Triangle facet $\mathbf{T}$}\label{sec:triangle_fact}
In our work, triangle facets are stored in regions. A triangle facet describes a small surface that exists in the reconstructed scene. It is maintained online by our \textit{meshing} module and is asynchronously copied to the \textit{broadcaster} module for publishing.  For a triangle facet $\mathbf{T}$, it is constituted by the following elements: 1) The sorted indices $\mathtt{Pts\_id}(\mathbf{T})$ of three mesh vertices that form this triangle: $ \mathtt{Pts\_id}(\mathbf{T}) = \{ i, j, k \},~i < j < k \label{eq_triangle_pts_id} $. 2) The center $\mathtt{Center}(\mathbf{T})$ and normal $\mathtt{Norm}(\mathbf{T})$ (both in the global reference frame) of this facet.
    
\subsubsection{Mesh vertex $\mathbf{V}$}\label{sec:mesh_vertices}
In ImMesh, mesh vertices are the points that constitute the geometric structure (shape) of mesh. For the $i$-th vertex $\mathbf{V}_i$, it contains the following elements: 1) The unique index (id) of this vertex $\mathtt{Id}(\mathbf{V}_i)$ in the global list containing all the vertices in the map. 2) Its 3D position $\mathtt{Pos}(\mathbf{V}_i)\in \mathbb{R}^3$ in the global frame. 3) The list $\mathtt{Tri\_list}(\mathbf{V}_i)$ of triangles facets whose vertices contain $\mathbf{V}_i$.

\subsection{Data structure: Hash tables and Incremental kd-Tree (ikd-Tree)}

In our work, we leverage a global list for accessing mesh vertices by indices. Besides, we employ two data structures (i.e., hash tables and incremental kd-tree (ikd-Tree)) for efficiently managing our four data types. Specifically, we leverage the hash tables for efficient lookup of regions, voxels, and triangle facets, and maintain an ikd-Tree to enable the fast kNN search of mesh vertices.

\subsubsection{Hash tables}\label{sect_3D_hash}
To facilitate efficient lookup of the data types (i.e., regions, voxels, and triangle facets), and avoid excessive memory consumption from allocating regular data structures in continuous memory space, we employ a spatial hashing scheme. This scheme allows us to compactly store, access, and update the data structure by mapping them into a hash table using appropriate hash functions, as illustrated in Fig. \ref{fig_hierarchical_voxels}.

Given a 3D vector $\mathbf{p} = [x,y,z]^T \in \mathbb{R}^3$, its corresponding hash key $\boldsymbol{\mathcal{H}}(\mathbf{p})$ is calculated via the 3D hash function $\mathtt{Hash}(x,y,z)$, shown as below:
\begin{align}
\boldsymbol{\mathcal{H}}(\mathbf{p}) &= \mathtt{Hash}(x,y,z) =  \mathtt{Int\_Hash}(x_i,y_i,z_i) \label{eq_hash_func} \\
&=  \mathtt{Mod}((x_i\cdot p_1) \oplus \HL{(y_i\cdot p_2)} \oplus (z_i \cdot p_3),n)   \\
\begin{split}
x_i = \mathtt{Round}&( x*100 / S ), \quad 
y_i = \mathtt{Round}( y*100 / S ) \\
&z_i = \mathtt{Round}( z*100 / S ) \label{eq_hash_func_interger}
\end{split}
\vspace{-0.2cm}
\end{align}
where $x_i,y_i,z_i$ are the corresponding integer-rounded coordinates, $S$ is size of a region (i.e., $S_{\mathbf{R}}$) or voxel (i.e., $S_{\mathbf{O}}$), $\oplus$ is the XOR operation, and function $\mathtt{Mod}(a, b)$ is the calculation of integer $a$ modulus another integer $b$. $p_1, p_2, p_3$ are three large prime numbers for reducing the collision probability \cite{teschner2003optimized, niessner2013real}, $n$ is the hash table size. In our work, we set the value of $p_1, p_2, p_3$ and $n$ as $116101, 37199, 93911$ and $201326611$, respectively. 

In our map structure, we maintain three independent hash tables for regions, voxels, and triangle facets, denoted as: $\boldsymbol{\Xi}_{\mathbf{R}}$, $\boldsymbol{\Xi}_{\mathbf{O}}$, and $\boldsymbol{\Xi}_{\mathbf{T}}$, respectively. For a region $\mathbf{R}_i$, a voxel  $\mathbf{O}_j$, and a triangle facet  $\mathbf{T}_k$, they are mapped to hash tables (i.e., $\boldsymbol{\Xi}_{\mathbf{R}}$, $\boldsymbol{\Xi}_{\mathbf{O}}$, and $\boldsymbol{\Xi}_{\mathbf{T}}$) through the hash keys $\boldsymbol{\mathcal{H}}_\mathbf{R}(\mathbf{R}_i)$, $\boldsymbol{\mathcal{H}}_\mathbf{O}(\mathbf{O}_j)$, and $\boldsymbol{\mathcal{H}}_\mathbf{T}(\mathbf{T}_k)$ are calculated as below:
\begin{align}
\mathbf{R} \mapsto \boldsymbol{\Xi}_{\mathbf{R}}: &~   \boldsymbol{\mathcal{H}}_\mathbf{R}(\mathbf{R}_i) =  \boldsymbol{\mathcal{H}}(\mathbf{p}_i),~~ \mathbf{p}_i \in \mathbb{R}^{3} \label{eq_hash_function_R}\\
\mathbf{O} \mapsto \boldsymbol{\Xi}_{\mathbf{O}}: &~
\boldsymbol{\mathcal{H}}_\mathbf{O}(\mathbf{O}_j) =  \boldsymbol{\mathcal{H}}(\mathbf{p}_j),~~ \mathbf{p}_j \in \mathbb{R}^{3} \label{eq_hash_function_O}\\
\begin{split}
\mathbf{T} \mapsto \boldsymbol{\Xi}_{\mathbf{T}}: &~
\boldsymbol{\mathcal{H}}_\mathbf{T}(\mathbf{T}_k) =  \mathtt{Int\_Hash}(\mathtt{Pts\_id}(\mathbf{T}_k)) 
\end{split}
\end{align}
where $\mathbf{p}_i$ (and $\mathbf{p}_j$) can be any point that located inside \HL{region} $\mathbf{R}_i$ (and voxel $\mathbf{O}_j$). The hash function $\boldsymbol{\mathcal{H}}_\mathbf{R}(\cdot)$ in (\ref{eq_hash_function_R}) and $\boldsymbol{\mathcal{H}}_\mathbf{R}(\cdot)$ in (\ref{eq_hash_function_O}) are distinguished with different container's size $S$ in (\ref{eq_hash_func_interger}).
 
Besides, we use function $\boldsymbol{\Psi}(\cdot)$  to denote the retrieval of $\mathbf{R}_i$, $\mathbf{O}_j$, and $\mathbf{T}_k$ from the hash tables, shown as follows:
\begin{align}
\mathbf{R} \mapsfrom \boldsymbol{\Xi}_{\mathbf{R}}: &~~
\mathbf{R}_i = \boldsymbol{\Psi}(\boldsymbol{\Xi}_{\mathbf{R}}, \boldsymbol{\mathcal{H}}_\mathbf{R}(\mathbf{R}_i) ) \label{eq_retrive_hash_function_R}\\
\mathbf{O} \mapsfrom \boldsymbol{\Xi}_{\mathbf{O}}: &~~
\mathbf{O}_j = \boldsymbol{\Psi}(\boldsymbol{\Xi}_{\mathbf{O}}, \boldsymbol{\mathcal{H}}_\mathbf{O}(\mathbf{O}_j) ) \label{eq_retrive_hash_function_O} \\
\mathbf{T} \mapsfrom \boldsymbol{\Xi}_{\mathbf{T}}: &~~
\mathbf{T}_k = \boldsymbol{\Psi}(\boldsymbol{\Xi}_{\mathbf{T}}, \boldsymbol{\mathcal{H}}_\mathbf{T}(\mathbf{T}_k) ) \label{eq_retrive_hash_function_K}
\end{align}
\indent Notice that the hash table is unstructured, indicating that neighboring regions (or voxels) are not stored spatially but in different parts of the buckets, as illustrated by two neighboring regions $\mathbf{R}_i$ and $\mathbf{R}_j$ in Fig.~\ref{fig_hierarchical_voxels}.

Lastly, for resolving the possible hash collision (i.e., two pieces of data in a hash table share the same hash value), we adopt the technique in \cite{niessner2013real}, using the implementation of $\mathtt{unordered\_map}$  container \cite{cppstdunorderedmap} in C++ standard library (std) \cite{ISO_Cpp_standard}.



\subsubsection{Incremental kd-Tree (ikd-Tree)}
\label{section_ikd_tree}
We maintain an incremental kd-tree to enable the fast kNN search of mesh vertices. The ikd-Tree is proposed in our previous work \cite{fast-lio2, ikd-tree}, which is an efficient dynamic space partition data structure for fast kNN search. Unlike existing static kd-trees (e.g., kd-tree implemented in PCL \cite{rusu20113d} and FLANN\cite{muja2009fast}) that require rebuilding the entire tree at each update, ikd-Tree achieves lower computation time by updating the tree with newly coming points in an incremental manner. In ImMesh, we use the ikd-Tree for: 1) ensuring that the distance between any two mesh vertices remains larger than the minimum value $\xi$, thereby maintaining the triangle mesh at a proper resolution. 2) enabling the vertex dilation operation in our voxel-wise meshing operation to erode the gaps between neighbor voxels.

\section{Receiver and localization}\label{sect_meth_receiver_locaization}
The \textit{receiver} module is designed for processing and packaging the input sensor data. As shown in the red box of Fig.~\ref{fig_overview}, our \textit{receiver} module receives the streaming of LiDAR data from live or offline recorded files, processes the data to a unified data format (i.e., customized point cloud data) that make ImMesh compatible with LiDARs of different manufacturers, scanning mechanisms (i.e., mechanical spinning, solid-state) and point cloud density (e.g., 64-, 32-, 16-lines, etc.). Besides, if the IMU source is available, our \textit{input} module will also package the IMU measurements within a LiDAR frame by referring to the sampling time.

The \textit{{localization}} module utilizes the input data stream of \textit{receiver} module, reuses the voxels for estimating the sensor poses of 6 DoF by registering the points to planes in voxels in real-time. Our \textit{{localization}} module is built upon our previous work VoxelMap \cite{voxelmap}, which represents the environment with the probabilistic planes and estimating pose with an iterated Kalman filter. 


\subsection{Voxel map construction}\label{sect_voxel_map_construction}
Our \textit{localization} is built by representing the environment with probabilistic planes, which accounts for both LiDAR measurement noises and sensor pose estimation errors, and constructs the voxel-volumetric maps in a coarse-to-fine adaptive resolution manner. Since the main focus of this work is on meshing, we only discuss those processes in \textit{localization} module that are closely related to our \textit{meshing} module. For the detailed modeling and analysis of LiDAR's measurement noise and sensor estimation errors, we recommend our readers to our previous work VoxelMap\cite{voxelmap}.

For each  LiDAR point, we first compensate the in-frame motion distortion with an IMU backward propagation introduced in \cite{fast-lio2}. Denoting ${}^L\mathbf{p}_i$ the $i$-th LiDAR point after motion compensation, it is registered to the world frame as ${}^W\mathbf{p}_i$ with the estimated sensor pose $({}^W_L {\mathbf R}, {}^W_L {\mathbf t} )\in SE(3)$:
\begin{equation}
{}^W \! \mathbf p_i = {}^W_L {\mathbf R} {}^L \mathbf p_i + {}^W_L {\mathbf t} \label{eq_point_lidar_to_world}
\end{equation}

The registered LiDAR point ${}^W \! \mathbf p_i$ is stored inside the voxels. Given all points ${}^W \! \mathbf p_i$ $(i=1,...,N)$ inside a voxel $\mathbf{O}$, the points covariance matrix $\mathbf{A}$ is 
\vspace{-0.2cm}
\begin{equation}
\label{mean_cov}
\bar{\mathbf p}=\frac{1}{N}\sum_{i=1}^N {}^W \! \mathbf p_i, \quad \mathbf A=\frac{1}{N}\textcolor{black}{\sum_{i=1}^N} \left({}^W \! \mathbf p_i -\bar{\mathbf p} \right) \left({}^W \! \mathbf p_i -\bar{\mathbf p} \right)^T
\end{equation}
where the symmetric matrix $\mathbf{A}$ depicted the distribution of pall oints. Perform the eigenvalue decomposition of matrix $\mathbf{A}$: 
\begin{align}
\hspace{-0.2cm}
\mathbf{A}\mathbf{U} &= 
\begin{bmatrix}
\lambda_1 & & \\
& \lambda_2 & \\
& & \lambda_3\\
\end{bmatrix}
\begin{bmatrix}
\mathbf{u}_1 & \mathbf{u}_2 & \mathbf{u}_3
\end{bmatrix}, ~~ \lambda_1\geq \lambda_2 \geq \lambda_3
 \label{eq_eigen_decompose_A}
\end{align}
where  $ \lambda_1, \lambda_2, \lambda_3 $ are the eigenvalues and $\mathbf{u}_1, \mathbf{u}_2 , \mathbf{u}_3$ are the correspondent eigenvectors.
In our \textit{meshing} module, we use these calculated eigenvectors of voxel $\mathbf{O}$ for performing the dimension reduction through projection, as we will discuss in Section \ref{sect_dimension_reduction}.

In our localization module, voxel $\mathbf{O}$ might be subdivided into smaller sub-voxels to construct possible planar features at finer resolutions for robust localization in unstructured environments. Then, the sensor pose $({}^W_L {\mathbf{R}}, {}^W_L {\mathbf{t}} )$ is estimated by minimizing the point-to-plane residual. While this paper primarily focuses on our mesh reconstruction method, we refer readers to our previous work \cite{voxelmap} for more details on the implementation of our \textit{localization} module, including the voxel subdivision and state estimation.

\subsection{Point cloud registration}\label{sect_point_cloud_registration}
With the estimated sensor pose $({}^W_L {\mathbf R}, {}^W_L {\mathbf t} )$, we perform the point cloud registration for transforming each measurement point ${}^L \mathbf p_i$ from the LiDAR frame to the global frame (i.e., the first LiDAR frame) with (\ref{eq_point_lidar_to_world}). This registered point cloud is then used for: 1) publishing to other applications with our \textit{broadcaster}. 2) updating the voxel map (detailed in \cite{voxelmap}). 3) appending to \textit{map structure} that serves as the mesh vertices for shaping the geometry structure of our online reconstructed triangle mesh.

 
If a new registered point does not lie on an existing voxel $\mathbf{O}$ (or region $\mathbf{R}$), a new voxel (or region) will be created and added to the hash table $\boldsymbol{\Xi}_{\mathbf{O}}$ (or $\boldsymbol{\Xi}_{\mathbf{R}}$). Subsequently, the newly registered point will be included in the newly constructed voxel.
 
\subsubsection{Append of mesh vertices}\label{sect_append_of_mesh_vertices} The registered LiDAR points are also used for forming the meshing vertices in \textit{map structure}. To be detailed,  we first leverage a voxel-grid filter to downsample the newly registered LiDAR point cloud. Then, to avoid the appearance of tiny triangles in reconstructing the mesh, we leverage the ikd-Tree for keeping the minimum distance $\xi$ between any of two meshing vertices. That is, for each register LiDAR point ${^W}\mathbf{p}_i$ in the global frame, we search for the nearest mesh vertex in \textit{map structure} with ikd-Tree. If the Euclidean distance between this point and the searched vertex is smaller than $\xi$, we will discard this point. Otherwise, this point will be used for: 1) constructing a new mesh vertex $\mathbf{V}_i$, where $i$ is the unique index indicating that $\mathbf{V}_i$ is the $i$-th appended vertex. 2) adding the vertex $\mathbf{V}_i$ to the ikd-Tree. 3) pushing back $\mathbf{V}_i$ to the vertex array of the voxel $\mathbf{O}_j$ that $\mathbf{V}_i$ lies in. After, the status flag $f_{\mathbf{O}_j}$ of  $\mathbf{O}_j$ is set as \textit{activated} for notifying the meshing module for performing the voxel-wise meshing operation.

\section{Meshing}\label{sect_meth_mesh}
In ImMesh, our meshing module takes the registered LiDAR scan for incrementally reconstructing the triangle mesh on the fly. We explicitly reconstruct the triangle mesh by directly utilizing 3D registered LiDAR points as mesh vertices enabled by two facts of LiDAR sensors: 1) The points sampled by LiDAR and registered via the LiDAR odometry and mapping \cite{voxelmap} have very high positional accuracy. Hence, they can accurately shape the geometric structure of the mesh.  2) A LiDAR measurement point naturally lies on the surface of the detected object, with two other points in the same plane that can form a triangle facet to represent its \HL{underlying} surface. 

\subsection{Goals and requirements}\label{sect_meshing_goad_requirements}
With the accurate mesh vertices appended from the point cloud registration in Section \ref{sect_point_cloud_registration}, the problem of online mesh reconstruction is converted to another goal, which is to seek a proper way of real-time reconstructing the triangle facets with a growing 3D point set. This new problem is barely researched to date. Given a set of growing 3D points, our \textit{{meshing}} module is designed to incrementally reconstruct the triangle facets considering the following four requirements: 

Firstly, precision is our primary consideration. For each reconstructed triangle facet representing the surface of the scene, we require it to lie on an existing plane.

Secondly, the reconstructed mesh should be hole-less. In the dense reconstruction of the surface triangle mesh, the appearance of holes is unacceptable since they lead to the wrong rendering results, where surfaces behind a real object are rendered. 

Thirdly, the reconstruction of triangle mesh should avoid constructing sliver triangles. A sliver triangle (i.e., the noodle-like triangle), as defined in the communities of computer graphics \cite{ stevens2002computer}, is a thin triangle whose area is nearly zero, an undesired property in the field of computer graphics. For example, these noodle-like triangles would cause some errors in the numerical analysis on them\cite {kahan1776miscalculating}. Besides, these unfavorable properties cause troubles in the pipelines of rendering (e.g., rasterization, texturing, and anti-aliasing\cite{woo1999opengl, laine2011high,  akenine2019real}), which leads to the loss of accuracy in calculating (e.g., depth testing, interpolation, etc.) the pixel values distributed near the sharp angle \cite{evans1996optimizing, hearn2004computer, akenine2019real}.

Lastly, the complexity of triangle mesh reconstruction should be computationally efficient to meet the requirement of real-time applications. The time consumption of each meshing process should not exceed the sampling duration of two consecutive LiDAR frames.   

\subsection{Challenges and approaches}
To achieve our goals of dense incremental meshing with the four requirements listed above, our system is proposed based on a deep analysis of the challenges. The challenges and corresponding scientific approaches are briefed below:
 
The first challenge is that the global map is continuously grown by the newly registered LiDAR points,  with each update of a LiDAR scan only affecting parts of the scene. Hence, an incremental mesh reconstruction method should be able to process only those parts of the scene with new points. In our work, we incrementally perform the mesh reconstruction with a mechanism similar to \textit{git}\cite{loeliger2012version}. For each incremental mesh update, we first retrieve the data of the voxels with new mesh vertices appended via the \textit{pull} step (detailed in Section~\ref{section_mesh_pull}). Then, an efficient voxel-wise meshing algorithm is executed to reconstruct the mesh with these data. The incremental modifications of newly reconstructed results w.r.t. pulled results are calculated in our \textit{commit} step (detailed in Section~\ref{sect_mesh_commit}). Finally, these incremental modifications are merged to the global map via our \textit{push} step (detailed in Section~\ref{sect_mesh_push}).

Given a set of 3D vertices, the second challenge is how to correctly and efficiently reconstruct the triangle facets representing the surfaces of the scene. Since it is hard to directly reconstruct mesh from these mesh vertices in 3D space, our work performs the meshing operation in 2D. To be detailed,  for vertices located in a voxel $\mathbf{O}$, we first project them into a proper plane (i.e., the estimated plane given by the \textit{{localization}} module). The mesh of these 2D points is constructed using the 2D meshing algorithms and is recovered back to 3D (detailed in Section~\ref{sect_2d_delaunay_triangulation}).

\subsection{Voxel-wise vertex retrieval}\label{section_pt_retrieving}
\subsubsection{Retrieval of in-voxel vertices}
To reconstruct the triangle mesh incrementally, the first step is to retrieve the vertices that need to mesh with the newly added points. ImMesh uses voxels for dividing the 3D space, and uses the flag $f_{\mathbf{O}}$ of each voxel $\mathbf{O}$ for identifying whether $\mathbf{O}$ has newly appended mesh vertices (i.e., \textit{activated} voxel). 

Take an \textit{activated} voxel $\mathbf{O}_i$ as an example. We perform a voxel-wise meshing operation to reconstruct the triangle facets with all in-voxel vertices. For all vertices inside the voxel $\mathbf{O}_i$, we denote them as $\boldsymbol{\mathcal{V}}_{i}^{\mathtt{In}} = \{\mathbf{V}_{j_1}, \mathbf{V}_{j_2}, ..., \mathbf{V}_{j_m} \}$.

\begin{figure}[t]
	\centering
        \setlength{\belowcaptionskip}{-0.3cm}
	\includegraphics[width=1.00\linewidth]{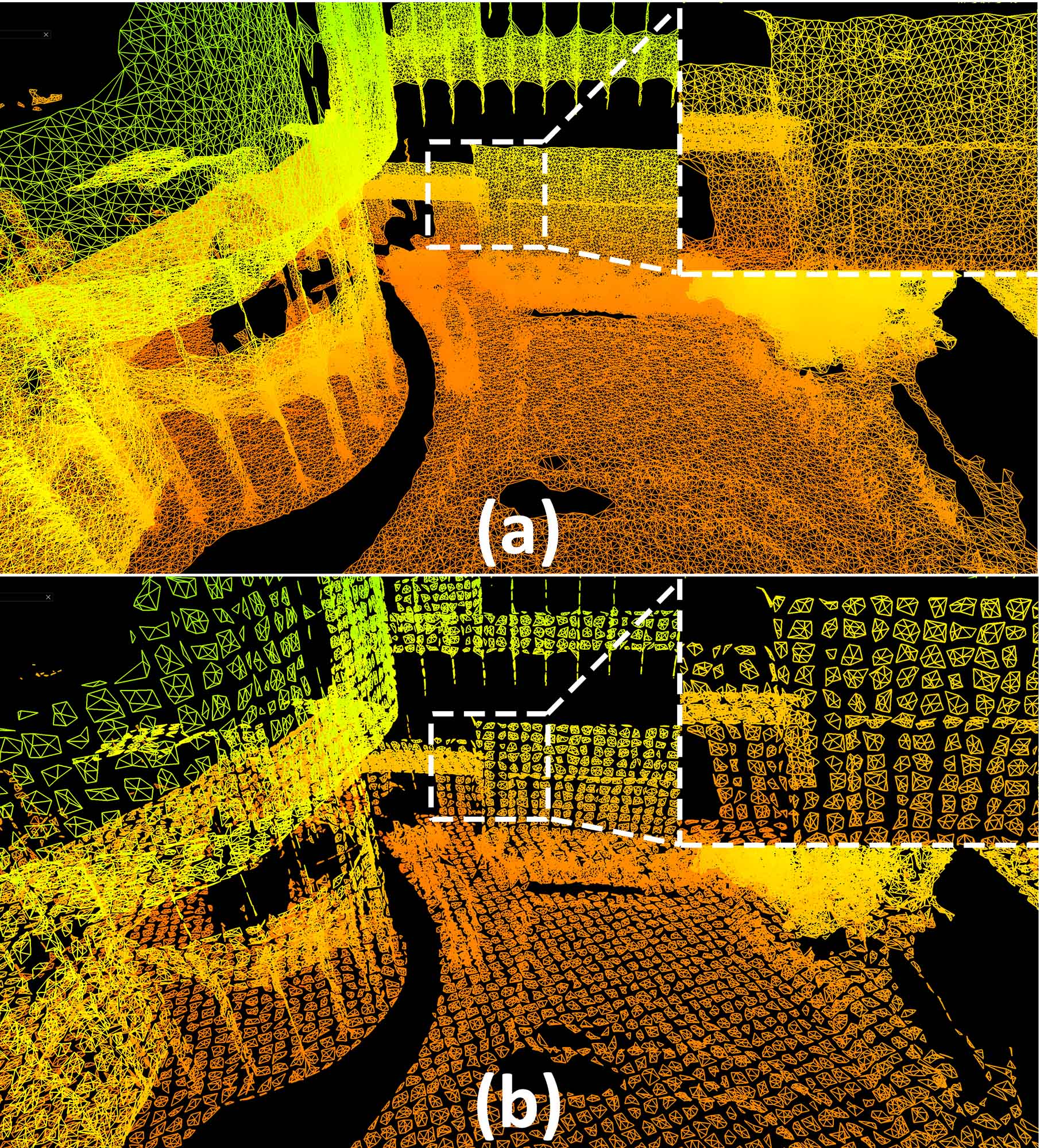}
	\caption{The comparisons of mesh reconstruction with (a) and without (b) the vertex dilation.}
	\label{fig_ikd_comp}
    \vspace{0.1cm}
\end{figure}

\subsubsection{Vertex dilation}
In practice, if we perform the meshing operation with only the in-voxel mesh vertices, the gaps between neighborhood voxels will appear due to the absence of triangles facets across voxels, as shown in Fig.~\ref{fig_ikd_comp}(b). Motivated by morphological operations (e.g., dilation and erosion) in digital image processing \cite{castleman1996digital}, we perform the 3D point cloud dilation for adding neighborhood points of $\boldsymbol{\mathcal{V}}_i^{\mathtt{In}}$ to erode the gaps between voxels, as shown in Fig.~\ref{fig_ikd_comp}(a).

For vertex $\mathbf{V}_{i_j} \in \boldsymbol{\mathcal{V}}_i^{\mathtt{In}}$, we perform the radius-search operation by leveraging the ikd-Tree \cite{ikd-tree} for searching the nearest vertices of $\mathbf{V}_{i_j}$ with their Euclidean distance smaller than a given value $d_r$ (usually set as $1/4$ of the size of a voxel). Using $\tilde{ \boldsymbol{\mathcal{V}}}_{i_j} $ to denote the searched neighbor vertices of $\mathbf{V}_{i_j}$, we have:
\begin{align}
	\forall \mathbf{V} \in  \tilde{ \boldsymbol{\mathcal{V}}}_{i_j}, \quad \left|| \mathtt{Pos}(\mathbf{V}) -  \mathtt{Pos}(\mathbf{V}_{i_j}) \right|| \leq d_r.
\end{align}

We enumerate each $\mathbf{V}_{i_j} \in \boldsymbol{\mathcal{V}}_i^{\mathtt{In}}$ and union the corresponding $\tilde{ \boldsymbol{\mathcal{V}}}_{i_j}$ into $\boldsymbol{\mathcal{V}}_i$ (excluding duplicated vertices), which is the set of dilated vertices. 
The full algorithm of our voxel-wise vertex retrieval is shown in Algorithm \ref{alg_voxel_wise_pt_retriving}.
\setlength{\textfloatsep}{-0.2cm}
\begin{algorithm}[h] 
	\small
	\caption{Voxel-wise vertex retrieval of $\mathbf{O}_i$ }
	\label{alg_voxel_wise_pt_retriving}
        \SetKwFunction{search}{RadiusSearch}
	\renewcommand{\thealgocf}{}
	\SetKwInOut{Input}{Input}
	\SetKwInOut{Output}{Output}
	\SetKwInOut{Begin}{Begin}
	\SetKwInOut{Start}{Start}
	\SetKwInOut{Return}{Return}
	\Input{The \textit{activated} voxel $\mathbf{O}_i$}
	\Output{The retrieved vertex set $\boldsymbol{\mathcal{V}}_i$ }
	\Start
	{Copy all in-voxel pointer list to $\boldsymbol{\mathcal{V}}_i^{\mathtt{In}}$. \\
    $\boldsymbol{\mathcal{V}}_i = \boldsymbol{\mathcal{V}}_i^{\mathtt{In}}$	.
	}
	\ForEach{$\mathbf{V}_{i_j} \in \boldsymbol{\mathcal{V}}_i^{\mathtt{In}} $}
	{
 $\tilde{ \boldsymbol{\mathcal{V}}}_{i_j}$ = \search{$\mathbf{V}_{i_j}$,$d_r$}\\
		\ForEach{ $ \mathbf{V} \in \tilde{ \boldsymbol{\mathcal{V}}}_{i_j}$ }	
		{
			\If{  $ \mathbf{V} \notin \boldsymbol{\mathcal{V}}_i $ }
			{
				$\boldsymbol{\mathcal{V}}_i = \boldsymbol{\mathcal{V}}_i \cup \mathbf{V}$
			}
		}
	}
	\Return{ The retrived vertex set $\boldsymbol{\mathcal{V}}_i$ after dilation }
	\vspace{-0.05cm}
\end{algorithm}
\setlength{\textfloatsep}{-0.0cm}

\subsection{Dimension reduction through  projection}\label{sect_dimension_reduction}
With the mesh vertices $\boldsymbol{\mathcal{V}}_i$ retrieved from Algorithm~\ref{alg_voxel_wise_pt_retriving}, we introduce the voxel-wise mesh reconstruction. 

\subsubsection{Projection of 3D vertices on a 2D plane}
 Since it is hard to directly mesh in real-time with $\boldsymbol{\mathcal{V}}_i$, which is distributed in 3D space, we simplify the 3D meshing problem into a 2D one by projecting $\boldsymbol{\mathcal{V}}_i$ on a suitable plane. This dimension reduction by projection is inspired by two key observations: 1) Every LiDAR point can be viewed as lying on a small local surface around it. Hence, for vertices $\boldsymbol{\mathcal{V}}_i$ retrieved from Algorithm~\ref{alg_voxel_wise_pt_retriving} that are distributed in a small area (i.e., inside a voxel $\mathbf{O}_i$), they tend to form a planar-like point cluster. 2) For these planar-like point clusters, we can approximately mesh them in a 2D view on their lying surface. 
To preserve the 3D space spanned by $\boldsymbol{\mathcal{V}}_i$ to the best extent, the plane $(\mathbf{n}, \mathbf{q})$ suitable for projection should be formed by the two principal components of $\boldsymbol{\mathcal{V}}_i$, which is essentially the plane fitted from $\boldsymbol{\mathcal{V}}_i$ and has already been calculated in our {\textit{localization}} module in Section \ref{sect_voxel_map_construction}. The norm $\mathbf{n}$ of the plane is the eigenvector $\mathbf{u}_3$ that corresponds to the minimum eigenvalue $\lambda_3$ in (\ref{eq_eigen_decompose_A}), which is the eigendecomposition of point covariance matrix $\mathbf{A}$ in voxel $\mathbf{O}_i$.  $\mathbf{q}$ is the center points inside  $\mathbf{O}_i$.


For each vertex $\mathbf{V}_{i_j}\in \boldsymbol{\mathcal{V}}_i $, we project it to plane $(\mathbf{n}, \mathbf{q})$. The resultant 2D point $\mathbf{p}_{i_j}$ is calculated as:
\vspace{-0.2cm}
\begin{align}
\mathbf{p}_{i_j} = \left[ \phi, \rho \right]^T& \in \mathbb{R}^{2} \\
\phi = \left( \mathtt{Pos}(\mathbf{V}_{i_j}) - \mathbf{q} \right)^T \mathbf{u}_1, ~~
\rho &= \left( \mathtt{Pos}(\mathbf{V}_{i_j}) - \mathbf{q} \right)^T \mathbf{u}_2
\end{align}
\vspace{-0.05cm}
\hspace{-0.17cm}where $\mathbf{u}_1, \mathbf{u}_2$ are the other two eigenvectors in (\ref{eq_eigen_decompose_A}). We use $\boldsymbol{\mathcal{P}}_i = \{ \mathbf{p}_{i_1},\mathbf{p}_{i_2},...,\mathbf{p}_{i_m} \}$ to denote the 2D point set after projected onto the plane.

\begin{figure}[t]
	\centering
	\includegraphics[width=1.0\linewidth]{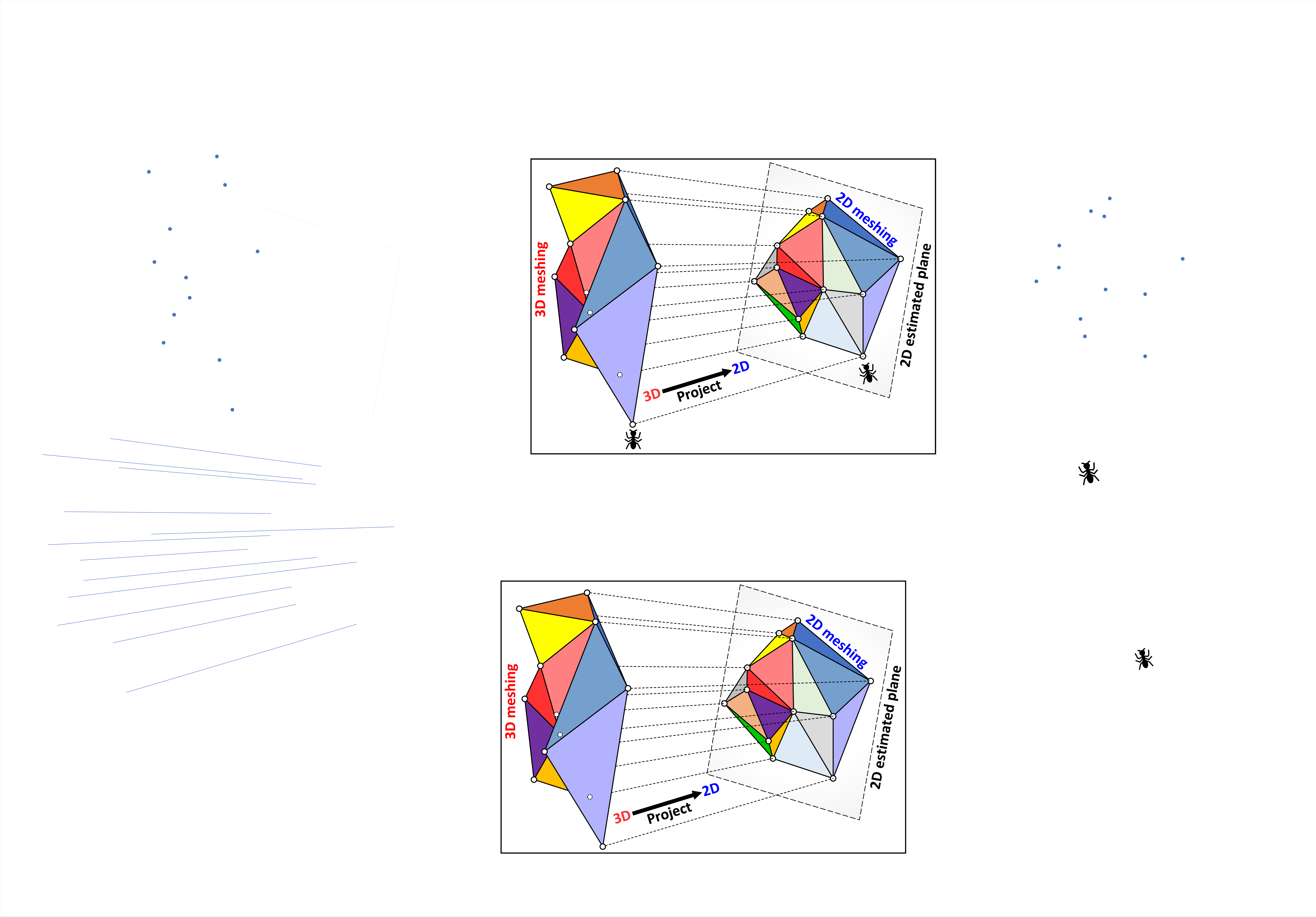}
	\caption{In ImMesh, we reduce the 3D meshing problem to a 2D one by projecting the 3D vertices onto their principal plane.}
	\label{fig_3D_2d_proj_del}
    \vspace{0.1cm}
\end{figure}

\subsubsection{Two-dimensional Delaunay triangulation}\label{sect_2d_delaunay_triangulation}
After the projection, the dimension of 3D meshing problem is reduced to a 2D one, which can be solved by 2D Delaunay triangulation.

As introduced in\cite{cgal, toth2017handbook}, a Delaunay triangulation $\mathbf{\mathtt{Del}(\boldsymbol{\mathcal{P}})}$ for a 2D point set $\boldsymbol{\mathcal{P}} = \{\mathbf{p}_{1}, \mathbf{p}_{2},..., \mathbf{p}_{m} \}$ is a triangulation such that no point in $\boldsymbol{\mathcal{P}}$ is inside the circumcircle of any triangle. Using  $\boldsymbol{\mathcal{T}}=\mathbf{\mathtt{Del}(\boldsymbol{\mathcal{P}})}$ to denote the triangle facets after triangulation, $\boldsymbol{\mathcal{T}}$ has the following properties: 1) Any of two facets are either disjoint or share a lower dimensional face (i.e., edge or point). 2) The set of facets in $\boldsymbol{\mathcal{T}}$ is connected with adjacency relation. 3) The domain $\mathbf{P}_{\boldsymbol{\mathcal{T}}}$, which is the union of facets in ${\boldsymbol{\mathcal{T}}}$, has no singularity\footnote{The union $\mathbf{U}_{\boldsymbol{\mathcal{T}}}$ of all simplices in $\boldsymbol{\mathcal{T}}$ is called the domain of $\boldsymbol{\mathcal{T}}$. A point in the domain of $\boldsymbol{\mathcal{T}}$ is said to be singular if its surrounding in $\mathbf{P}_{\boldsymbol{\mathcal{T}}}$ is neither a topological ball nor a topological disc (view \href{https://doc.cgal.org/latest/Triangulation_2/index.html}{\url{https://doc.cgal.org/latest/Triangulation\_2/index.html}} of \cite{cgal} for detail).}. With these three useful properties, the 2D Delaunay triangulation has been widely applied for reconstructing dense facets with a given 2D point set (e.g., \cite{rosinol2020kimera}).

Considering our requirements in Section \ref{sect_meshing_goad_requirements}, we chose Delaunay triangulation to reconstruct the mesh for its remarkable properties as follows. Firstly, it is a 2D triangulation providing mesh with no hole left in the convex hull of $\boldsymbol{\mathcal{P}}$, which satisfies our first requirement. Secondly, it naturally avoids sliver triangles by maximizing the minimum angles of the triangles in triangulation, which meets our second requirement. Finally, it is a fast algorithm suitable for real-time requirements. The algorithm complexity of $n$ points is $\boldsymbol{\mathcal{O}}(n\mathtt{log}(n))$ in 2D (p.s. $\boldsymbol{\mathcal{O}}(n^2)$ in 3D) \cite{attali2003complexity}.


Denote the triangle facets after the Delaunay triangulation of $\boldsymbol{\mathcal{P}}_i$ (from Section~\ref{sect_dimension_reduction}) as  $\boldsymbol{\mathcal{T}}_i = \mathbf{\mathtt{Del}}(\boldsymbol{\mathcal{P}}_i) =  \{\mathbf{T}_{i_1}, \mathbf{T}_{i_2},..., \mathbf{T}_{i_n} \}$. For each triangle facets $\mathbf{T}_{i_j} \in \boldsymbol{\mathcal{T}}_i$, we retrieve the indices of its three vertices with: $\{\alpha, \beta, \gamma\} = \mathbf{\mathtt{Pts\_id}}(\mathbf{T}_{i_j})$, indicating that this triangle is formed with 2D points $\{ \mathbf{p}_{i_\alpha}, \mathbf{p}_{i_\beta}, \mathbf{p}_{i_\gamma} \}$. Returning back to 3D space, we constitute a triangle facet $\mathbf{T}_{i_j}$ with vertices  $\{ \mathbf{V}_{i_\alpha}, \mathbf{V}_{i_\beta}, \mathbf{V}_{i_\gamma} \}$, as shown in Fig.~\ref{fig_3D_2d_proj_del}. Then, the center $\mathtt{Center}(\mathbf{T}_{i_j})$ and norm $\mathtt{Norm}(\mathbf{T}_{i_j})$ of $\mathbf{T}_{i_j}$ are calculated as below:
\begin{small}
\begin{align}
 \mathtt{Center}(\mathbf{T}_{i_j}) &= \left( \mathtt{Pos}(\mathbf{V}_{i_\alpha}) + \mathtt{Pos}(\mathbf{V}_{i_\beta}) + \mathtt{Pos}(\mathbf{V}_{i_\gamma}) \right)/3 \label{eq_triangle_center_tij} \\
                    \mathtt{Norm}(\mathbf{T}_{i_j}) &= \mathbf{n} / (||\mathbf{n}||) \label{eq_triangle_norm_tij} \\
              \mathbf{n} = ( \mathtt{Pos}(\mathbf{V}_{i_\alpha}) & - \mathtt{Pos}(\mathbf{V}_{i_\beta})  )  \times  \left( \mathtt{Pos}(\mathbf{V}_{i_\gamma}) - \mathtt{Pos}(\mathbf{V}_{i_\beta}) \right)
\vspace{-0.1cm}
\end{align}
\end{small}
\indent Additionally, to ensure proper face orientation for identifying the front-back face, which is crucial for various computer graphics applications such as front-back face culling, lighting, and shading, we adjust the normal of $\mathbf{T}_{i_j}$ to make it always face towards the current LiDAR position by:
\begin{align}
    \mathtt{If}:& \quad ( ({}^W_L{\mathbf t} - \mathtt{Center}(\mathbf{T}_{i_j}) )^T \mathtt{Norm}(\mathbf{T}_{i_j})  < 0~ \\
     \mathtt{Then}:& \quad \mathtt{Norm}(\mathbf{T}_{i_j}) = - \mathtt{Norm}(\mathbf{T}_{i_j}) \label{eq_flipped_normal}
\end{align}
where ${}^W_L{\mathbf t}$ is the LiDAR position of current scan, \HL{which} is estimated in our \textit{localization} module. \HL{Furthermore, if the normal is flipped in (\ref{eq_flipped_normal}), we will change the indices of $\mathbf{T}_{i_j}$ from $\{\alpha, \beta,\gamma\}$ to $\{\beta, \alpha, \gamma\}$ when publishing this facet in our \textit{broadcaster}, which is necessary to ensure the correct normal orientation in certain rendering engines (e.g., in \cite{oglfaceculling}).}



\subsection{Voxel-wise meshing with pull, commit, and push}\label{sect_meth_vx_mesh_pull_commit_push}
With the triangle facets $\boldsymbol{\mathcal{T}}_i$ newly constructed by the voxel-wise meshing operation, we incrementally merge $\boldsymbol{\mathcal{T}}_i$ to the existing triangle facets in the voxel currently saved in \textit{map structure}. This update is designed with a mechanism similar to \textit{git}\cite{loeliger2012version} (a version control software) that includes \textit{pull}, \textit{commit}, and \textit{push} steps.
\subsubsection{Pull}\label{section_mesh_pull} The pull operation aims to retrieve existing triangle facets $\boldsymbol{\mathcal{T}}^{\mathtt{Pull}}_i$ in the $i$-th $L2$ voxel. Given vertices $\boldsymbol{\mathcal{V}}_i$ in the voxel, which is obtained from Algorithm \ref{alg_voxel_wise_pt_retriving}, we retrieve the triangle facets $\boldsymbol{\mathcal{T}}^{\mathtt{Pull}}_i$ from the \textit{map structure} as shown in Algorithm \ref{alg_voxel_wise_mesh_pull}.
\setlength{\textfloatsep}{-0.3cm}
\begin{algorithm}[h]
	\small 
	\caption{Voxel-wise mesh pull.}
	\label{alg_voxel_wise_mesh_pull}
	\renewcommand{\thealgocf}{}
	\SetKwInOut{Input}{Input}
	\SetKwInOut{Output}{Output}
	\SetKwInOut{Begin}{Begin}
	\SetKwInOut{Start}{Start}
	\SetKwInOut{Return}{Return}
	\Input{The retrieved vertex set $\boldsymbol{\mathcal{V}}_i$ from Algorithm \ref{alg_voxel_wise_pt_retriving}} 
	\Output{Existing triangles facets in the voxel $\boldsymbol{\mathcal{T}}^{\mathtt{Pull}}_i$}
	\Start
	{$\boldsymbol{\mathcal{T}}^{\mathtt{Pull}}_i =  \{ \mathtt{null} \}$ 	
	}
	\ForEach{$\mathbf{V}_j \in \boldsymbol{\mathcal{V}}_i$}
	{
		Get triangles having vertex $\mathbf{V}_j$:	$\boldsymbol{\mathcal{T}}_{\mathbf{V}_j} = \mathbf{\mathtt{Tri\_List}}(\mathbf{V}_j)$
		\ForEach{$\mathbf{T}_k \in \boldsymbol{\mathcal{T}}_{\mathbf{V}_j}$}
		{
			Get all vertices of $\mathbf{T}_k$: $\{ \alpha, \beta, \gamma \} = \mathtt{Pts\_id}(\mathbf{T}_k)$
			
			\If{$(\mathbf{V}_\alpha \in \boldsymbol{\mathcal{V}}_i)$ and $(\mathbf{V}_\beta \in \boldsymbol{\mathcal{V}}_i)$ and $(\mathbf{V}_\gamma \in \boldsymbol{\mathcal{V}}_i)$ }
			{
				$\boldsymbol{\mathcal{T}}^{\mathtt{Pull}}_i = \boldsymbol{\mathcal{T}}^{\mathtt{Pull}}_i \cup \mathbf{T}_k $ 
			}
		}		
	}
	\Return{$\boldsymbol{\mathcal{T}}^{\mathtt{Pull}}_i$ }
	\vspace{-0.02cm}
\end{algorithm}
	
\subsubsection{Commit}\label{sect_mesh_commit} 
In this step, we incrementally update the newly reconstructed triangle facets $\boldsymbol{\mathcal{T}}_i$ (in Section \ref{sect_2d_delaunay_triangulation}) to the existing facets $\boldsymbol{\mathcal{T}}^{\mathtt{Pull}}_i$ (from Algorithm~\ref{alg_voxel_wise_mesh_pull}). These incremental updates are summarized into an array of mesh facets to be added $\boldsymbol{\mathcal{T}}^{\mathtt{Add}}_i$ and an array of mesh facets to be erased $\boldsymbol{\mathcal{T}}^{\mathtt{Erase}}_i$. The detailed processes of this commit step are shown in Algorithm \ref{alg_voxel_wise_mesh_commit}.
\begin{algorithm}[h]\small
	\caption{Voxel-wise mesh commit.}
	\label{alg_voxel_wise_mesh_commit}
	\renewcommand{\thealgocf}{}
	\SetKwInOut{Input}{Input}
	\SetKwInOut{Output}{Output}
	\SetKwInOut{Begin}{Begin}
	\SetKwInOut{Start}{Start}
	\SetKwInOut{Return}{Return}
	\Input{The pulled triangle facets $\boldsymbol{\mathcal{T}}^{\mathtt{Pull}}_i$ from Algorithm \ref{alg_voxel_wise_mesh_pull} \\ 
		   The reconstructed triangle facets $\boldsymbol{\mathcal{T}}_i$}
	\Output{The triangle facets to be added $\boldsymbol{\mathcal{T}}^{\mathtt{Add}}_i$.\\
		The triangle facets to be erased $\boldsymbol{\mathcal{T}}^{\mathtt{Erase}}_i$.\\  }
	\Start
	{$\boldsymbol{\mathcal{T}}^{\mathtt{Add}}_i =  \{ \mathtt{null} \}, \quad \boldsymbol{\mathcal{T}}^{\mathtt{Erase}}_i =  \{ \mathtt{null} \} $	
	}
	\ForEach{$\mathbf{T}_j \in \boldsymbol{\mathcal{T}}_i$}
	{
		\If{ $\mathbf{T}_j \notin \boldsymbol{\mathcal{T}}^{\mathtt{Pull}}_i$ }
		{
			$\boldsymbol{\mathcal{T}}^{\mathtt{Add}}_i = \boldsymbol{\mathcal{T}}^{\mathtt{Add}}_i \cup \mathbf{T}_j $
		}
	}
	\ForEach{$\mathbf{T}_j \in \boldsymbol{\mathcal{T}}^{\mathtt{Pull}}_i$}
	{
		\If{ $\mathbf{T}_j \notin \boldsymbol{\mathcal{T}}_i$ }
		{
			$\boldsymbol{\mathcal{T}}^{\mathtt{Erase}}_i = \boldsymbol{\mathcal{T}}^{\mathtt{Erase}}_i \cup \mathbf{T}_j $
		}
	}
	\Return{ The triangle facets to be added $\boldsymbol{\mathcal{T}}^{\mathtt{Add}}_i$ and erased $\boldsymbol{\mathcal{T}}^{\mathtt{Erase}}_i$. }
	\vspace{-0.02cm}
\end{algorithm}


\setlength{\textfloatsep}{-0.1cm}
\begin{algorithm}[t]
    \small
    \caption{Voxel-wise mesh push.}
    \label{alg_voxel_wise_mesh_push}
    \renewcommand{\thealgocf}{}
    \SetKwInOut{Input}{Input}
    \SetKwInOut{Output}{Output}
    \SetKwInOut{Begin}{Begin}
    \SetKwInOut{Start}{Start}
    \SetKwInOut{Return}{Return}
    \Input{The triangle facets that need to erased $\boldsymbol{\mathcal{T}}^{\mathtt{Erase}}_i$.\\
        The triangle facets that need to added $\boldsymbol{\mathcal{T}}^{\mathtt{Add}}_i$. }
    \SetKwProg{Fn}{Function}{:}{}
    \SetKwFunction{FTriAdd}{Add\_triangle}
    \SetKwFunction{FTriErasion}{Erase\_triangle}
    \Fn{\FTriAdd{$\mathbf{T}_j$}}
    {
        Get vertex indices $\{ \alpha, \beta, \gamma \} = \mathtt{Pts\_id}(\mathbf{T}_j) $\\
        Find the region $\mathbf{R}_k$ with $\mathtt{Center}(\mathbf{T}_j)$ via (\ref{eq_retrive_hash_function_R}).\\
        Set the status flag $f_{\mathbf{R}_k}$ of region $\mathbf{R}_k$ to \textit{Sync-required}. \\
        Add $\mathbf{T}^{G}_j$ to region $\mathbf{R}_k$ and triangles list of vertices $\mathbf{V}_{\alpha}$, $\mathbf{V}_{\beta}$, $\mathbf{V}_{\gamma}$
    }
    \Fn{\FTriErasion{$\mathbf{T}_j$}}
    {
        Get vertex indices $\{ \alpha, \beta, \gamma \} = \mathtt{Pts\_id}(\mathbf{T}_j) $\\
        Remove $\mathbf{T}_j$ from triangles list of vertices $\mathbf{V}_{\alpha}$, $\mathbf{V}_{\beta}$, $\mathbf{V}_{\gamma}$.\\
        Find the region $\mathbf{R}_k$ with $\mathtt{Center}(\mathbf{T}_j)$ via (\ref{eq_retrive_hash_function_R}).\\
        Set the status flag $f_{\mathbf{R}_k}$ of region $\mathbf{R}_k$ to \textit{Sync-required}. \\
        Remove $\mathbf{T}_j$ from region $\mathbf{R}_k$.\\
        Delete triangle $\mathbf{T}_j $ from memory. \\
    }
    \ForEach{$\mathbf{T}_j \in \boldsymbol{\mathcal{T}}^{\mathtt{Add}}_i$}
    {
        \FTriAdd{$\mathbf{T}_j$}
    }
    \ForEach{$\mathbf{T}_j \in \boldsymbol{\mathcal{T}}^{\mathtt{Erase}}_i$}
    {
        \FTriErasion{$\mathbf{T}_j$}
    }
\end{algorithm}

\subsubsection{Push}\label{sect_mesh_push} 
With the incremental modification $\boldsymbol{\mathcal{T}}^{\mathtt{Erase}}_i$ and $\boldsymbol{\mathcal{T}}^{\mathtt{Add}}_i$ from the previous \textit{commit} step, we perform the addition and erasion operations of triangle facets in \textit{push} step by: 1) constructing (or deleting) the triangle facet structures (as defined in Section~\ref{sec:triangle_fact}). 2) adding (or removing) the pointer to these facet structures to other data structures (i.e.,  mesh vertices and regions). The detailed processes of \textit{push} step are shown in Algorithm \ref{alg_voxel_wise_mesh_push}.

\subsection{Parallelism}\label{sect_parallelism}
To further improve the real-time performance, we implement our algorithms with parallelism for better utilization of the computation power of a multi-core CPU. In ImMesh, we have two major parallelisms as follows:

The first parallelism is implemented between the \textit{localization} module and the \textit{meshing} module. Except for the point cloud registration in {\textit{localization}} module, which needs to operate the mesh vertices as the meshing operation, the remaining processes of \textit{localization} module are parallelized with the \textit{meshing} module. More specifically, once our meshing processes start, the {\textit{localization}} module is allowed to process the new incoming LiDAR scans for estimation of the pose of LiDAR. However, the subsequent point cloud registration step is only allowed to be executed after the end of the current meshing process.

The second parallelism is implemented among the voxel-wise meshing operation of each \textit{activated} voxel. The voxel-wise meshing operations of different voxels are independent; thus, no conflicted operations exist on the same set of data. 

\subsection{The full meshing algorithm}\label{sect_full_mesh_alg}
To sum up, our full meshing processes are shown in Algorithm  \ref{alg_voxel_wise_full}.
\vspace{-0.3cm}
\setlength{\textfloatsep}{-0.1cm}
\begin{algorithm}[h]\small
	\caption{The full meshing process of each update of LiDAR scan}
	\label{alg_voxel_wise_full}
	\renewcommand{\thealgocf}{}
	\SetKwInOut{Input}{Input}
	\SetKwInOut{Output}{Output}
	\SetKwInOut{Begin}{Begin}
	\SetKwInOut{Start}{Start}
	\SetKwInOut{Return}{Return}
    \SetKwInOut{End}{End}
	\SetKwFunction{FTriAdd}{Add\_triangle}
	\SetKwFunction{FTriErasion}{Erase\_triangle}
	\SetKwFor{ForAllInPara}{foreach}{do in parallel}{end}
	\Input{The set of voxels $\boldsymbol{\mathcal{O}} = \{\mathbf{O}_1,\mathbf{O}_2,...,\mathbf{O}_m\}$ that \textit{activated} in Section \ref{sect_point_cloud_registration}}
	\Start
	{The triangle facets that need to added $\boldsymbol{\mathcal{T}}^{\mathtt{Add}}=\{\mathtt{null}\}$,
		and to be erased $\boldsymbol{\mathcal{T}}^{\mathtt{Erase}}=\{\mathtt{null}\}$ in this update .
	}
	\ForAllInPara{$\mathbf{O}_i \in \boldsymbol{\mathcal{O}}$}
	{
		Retrieve vertices $\boldsymbol{\mathcal{V}}_i$ with Algorithm \ref{alg_voxel_wise_pt_retriving}.\\
		Reconstruct the triangle facets $\boldsymbol{\mathcal{T}}_i$ with $\boldsymbol{\mathcal{V}}_i$ (Section \ref{sect_2d_delaunay_triangulation}),\\
		Performing voxel-wise mesh \textit{pull} (Algorithm \ref{alg_voxel_wise_mesh_pull}) to get $\boldsymbol{\mathcal{T}}^\mathtt{Pull}_i$. \Comment{// Mesh \textit{pull}~~~~~}\\
		Performing voxel-wise mesh \textit{commit} (Algorithm \ref{alg_voxel_wise_mesh_commit}) to get the triangle facets that need to be added $\boldsymbol{\mathcal{T}}^{\mathtt{Add}}_i$ and erased $\boldsymbol{\mathcal{T}}^{\mathtt{Erase}}_i$. \Comment{// Mesh \textit{commit}}\\
		$\boldsymbol{\mathcal{T}}^{\mathtt{Add}} = \boldsymbol{\mathcal{T}}^{\mathtt{Add}} \bigcup \boldsymbol{\mathcal{T}}^{\mathtt{Add}}_i $, $\quad$
		$\boldsymbol{\mathcal{T}}^{\mathtt{Erase}} = \boldsymbol{\mathcal{T}}^{\mathtt{Erase}} \bigcup \boldsymbol{\mathcal{T}}^{\mathtt{Erase}}_i $
	}
    \tcc{=== Mesh \textit{push} start ===}
	\ForEach{$\mathbf{T}_j \in \boldsymbol{\mathcal{T}}^{\mathtt{Add}}$}
	{
		\FTriAdd{$\mathbf{T}_j$} \Comment{// In Algorithm \ref{alg_voxel_wise_mesh_push}}
	}
	\ForEach{$\mathbf{T}_j \in \boldsymbol{\mathcal{T}}^{\mathtt{Erase}}$}
	{
		\FTriErasion{$\mathbf{T}_j$} \Comment{// In Algorithm \ref{alg_voxel_wise_mesh_push}}
	}
    \tcc{=== Mesh \textit{push} end ===}
    \ForEach{$\mathbf{O}_i \in \boldsymbol{\mathcal{O}}$}
    {
        Reset status flag $f_{\mathbf{O}_i}$ of $\mathbf{O}_i$ as \textit{deactived}.
    }
    \footnotesize
\end{algorithm}
\setlength{\textfloatsep}{-0.0cm}

\section{Broadcaster}\label{sect_meth_broadcaster}
In ImMesh, the \textit{broadcaster} module publishes our state estimation results (i.e., odometry) and mapping results (i.e., newly registered point cloud and triangle mesh) to other applications. Additionally, if a depth image is needed, the \textit{broadcaster} module will rasterize the triangle meshes into a depth image.

\subsection{Broadcast of triangle facets}\label{sect_refresh_triangles}
Since the triangle facets are stored in regions in an unstructured way, they can not be directly applied for broadcast. To resolve this problem, our {\textit{broadcaster}} module maintains a background thread that asynchronously copies the triangle facets from each \textit{sync-required} region (set as \textit{sync-required} after the triangle facets are updated in Algorithm \ref{alg_voxel_wise_mesh_push}) to a structured array for broadcasting. Then, these \textit {sync-required} regions are marked as \textit{synced} after the copying.
Finally, The \textit{{broadcaster}} module publishes the newest triangle facets to other applications.

\subsection{Rasterization of depth image}\label{sect_rasterization_depth_image}
Some robotic applications, such as autonomous navigation \cite{zhou2021raptor} and exploration\cite{zhou2021fuel} tasks, require dense accurate depth images for obstacle avoidance. To meet the requirements of these scenarios, the broadcaster module utilizes the triangle facets from Section \ref{sect_refresh_triangles} to rasterize a depth image at any customized resolution and FoV, based on the fast implementation of \textit{OpenGL}\cite{woo1999opengl}.

Besides depth image rasterization, the mesh obtained by our meshing module can reinforce the raw LiDAR point cloud measurements by increasing the resolution and enlarging the FoV. In detail, with the projection matrix and estimated pose used for rasterizing the depth image, the 3D points are obtained (i.e., unproject) from each pixel of the depth image. The unprojected 3D points would have higher resolution and larger FoV than the raw LiDAR measurement scan (see our Application-1 in Section~\ref{Application_1_LiDAR_reinforce}).

\iftrue

\section{Experiments and results}\label{sect_experiments_and_results}

In this paper, we conduct the experiments by evaluating our meshing ability, especially on the runtime performance and accuracy in reconstructing the triangle mesh.  

\subsection{Experiment-1: ImMesh for immediate mesh reconstruction}\label{sect_experiment_1}
In this experiment, we verify the overall performance of ImMesh toward real-time simultaneous localization and meshing with live video demonstrations. As shown in Fig.~\ref{fig_experiment_1}(b), we record the entire process of our data collection at the campus of the University of Hong Kong (HKU), deploying the ImMesh for simultaneously estimating the sensor pose and reconstructing the triangle mesh on the fly. The  \href{https://youtu.be/pzT2fMwz428?t=9}{\VIDEO{accompanying video \cite{immeshvideo} (starting at 00:09)}} demonstration of this experiment is available on YouTube.

\begin{figure}[t]
	\centering
       \setlength{\belowcaptionskip}{-0.6cm}
	\includegraphics[width=1.0\linewidth]{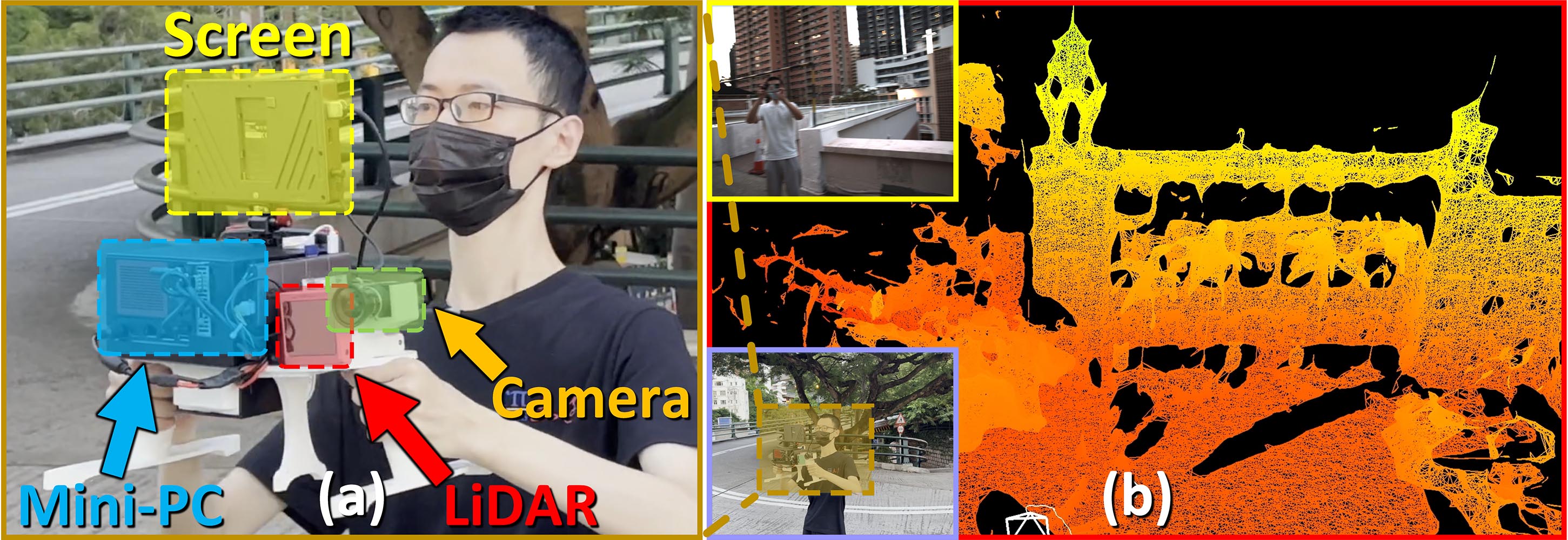}
	\caption{(a) shows our handheld device for data collection and online mesh reconstruction. (b) shows a snapshot of our \href{https://youtu.be/pzT2fMwz428?t=9}{\VIDEO{accompanying video \cite{immeshvideo} (starting at 00:09)}} of Experiment-1, with three time-aligned views of different sources including a screen-recorded view (in red), a camera preview (in yellow), and a third-person view (in blue). }
    \label{fig_experiment_1}
    \vspace{-1cm}
\end{figure}

\subsubsection{Experiment setup}\label{sect_experimetn_setup}
Our handheld device for data collection is shown in Fig.~\ref{fig_experiment_1}(a), which includes a mini-computer (equipped with an \textit{Intel i9-10900} CPU and 64 GB RAM), a \textit{Livox avia} 3D LiDAR (FoV: \SI{70.4}{\degree\times}\SI{77.2}{\degree}), and an RGB camera for previewing. In this experiment video, three time-aligned views of different sources are presented, including: 1) a screen-recorded view that shows the estimated pose and online reconstructed triangles mesh of ImMesh. 2) a camera preview that records the video stream of the front-facing camera. 3) a third-person view that records the whole process of this experiment.

\begin{table*}[h]
    \centering
    \begin{threeparttable}
\caption{The specifications of LiDARs in four datasets}
\label{table_lidar_spec}
\small
\setlength\tabcolsep{11.0pt}
\renewcommand{\arraystretch}{0.95} 
\begin{tabular}{ccccc}
\toprule
\textbf{Dataset} & \textbf{Kitti}                                                           & \textbf{NCLT}                                                            & \textbf{NTU VIRAL}                                                       & \textbf{R$^3$LIVE}                                                 \\ \toprule
 \multirow{2}{*}{LiDAR}	  &
\begin{minipage}{0.15\textwidth}
	\centering
      \includegraphics[height=15mm]{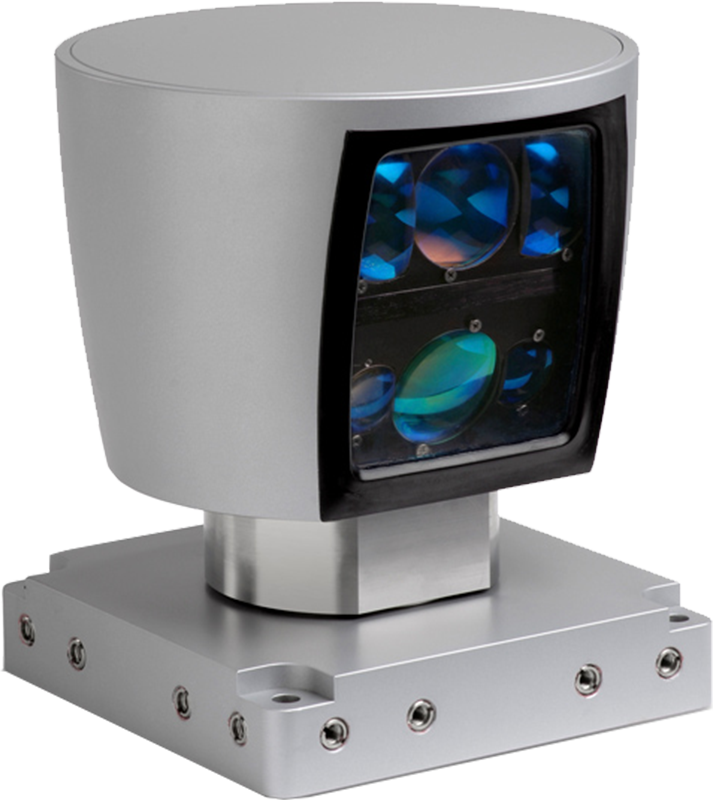}
\end{minipage} 
&   
\begin{minipage}{0.15\textwidth}
	\centering
      \includegraphics[height=14mm]{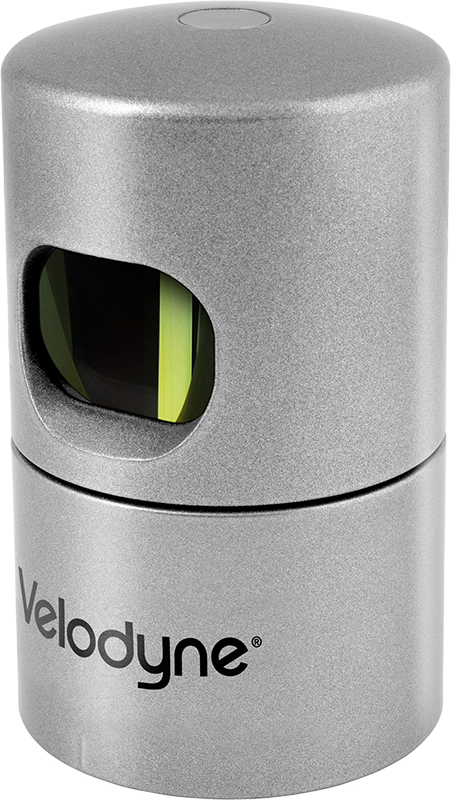}
\end{minipage}           
&     
\begin{minipage}{0.15\textwidth}
	\centering
      \includegraphics[height=15mm]{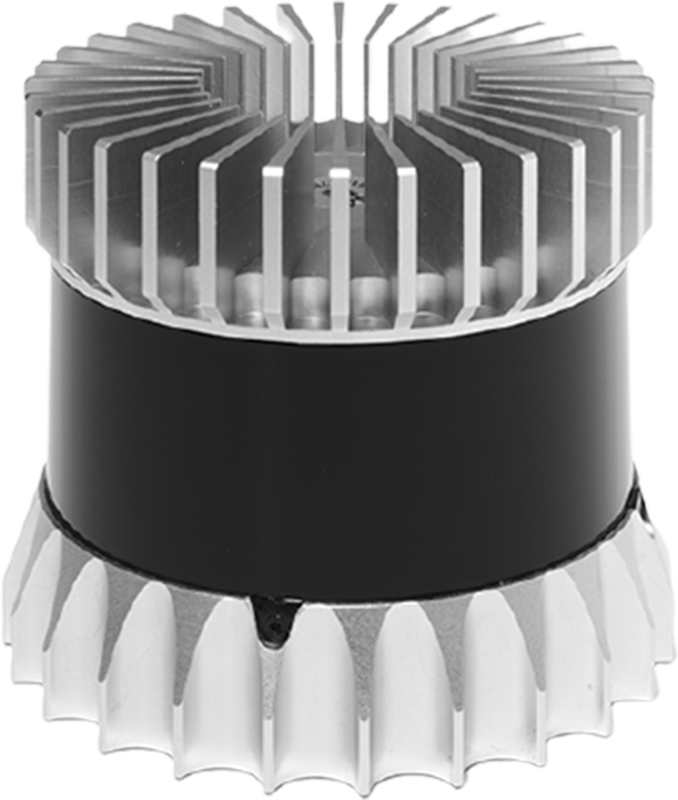}
\end{minipage}
&
\begin{minipage}{0.15\textwidth}
	\centering
      \includegraphics[height=14mm]{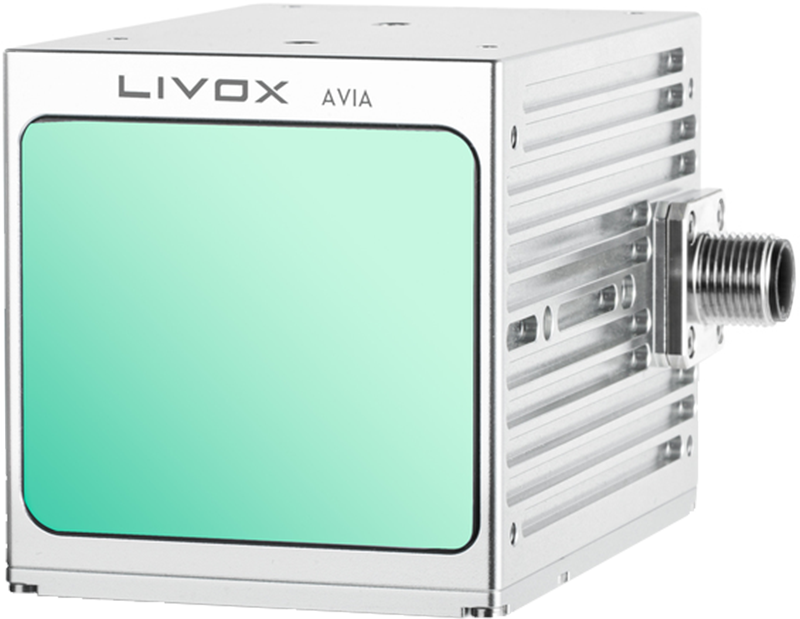}
\end{minipage}
\\ 
 \multirow{3}{*}{ }                                        & Velodyne HDL-64E                                                         & Velodyne HDL-32E                                                         & Ouster OS1-16 Gen1                                                       & Livox Avia                                                              \\   \toprule 
\begin{tabular}[c]{@{}c@{}}Scanning   \\      mechanism\end{tabular} & \begin{tabular}[c]{@{}c@{}}Mechanical,\\ spinning 64-line \end{tabular}          & \begin{tabular}[c]{@{}c@{}}Mechanical,\\      spinning 32-line\end{tabular}          & \begin{tabular}[c]{@{}c@{}}Mechanical,\\ spinning 16-line\end{tabular}           & \begin{tabular}[c]{@{}c@{}}Solid-state,\\      Risley prism\end{tabular}   \\ \toprule
\begin{tabular}[c]{@{}c@{}}Field of View\\ (Horizontal$^\circ$ $\times$ Vertical$^\circ$)\end{tabular} & $360.0^\circ \times 26.8^\circ$                                                    & $360.0^\circ \times 41.3^\circ$                                                    & $360.0^\circ \times 33.2^\circ$                                                     & $70.4^\circ \times 77.2^\circ$                                                            \\ \toprule
\begin{tabular}[c]{@{}c@{}}Points per second$^{[1]}$\end{tabular} &  1,333,312  & 695,000 & 327,680 & 240,000 \\ \toprule
\begin{tabular}[c]{@{}c@{}}Price (U.S. Dollar)\end{tabular} & \$~75,000 & \$~8,800 & \$~3,500 & \$~1,599
\\ \toprule
\end{tabular}
\vspace{-0.1cm}
\begin{tablenotes}
	\small
	\item[1] Only show the point rate of single-return mode.
\end{tablenotes}
\end{threeparttable}
\vspace{0.2cm}

\caption{This table shows the detailed information (e.g., length, duration, scenarios) of each testing sequence, the time consumption of ImMesh in processing a LiDAR scan, and the number of vertices and facets of each reconstructed mesh in Experiment-2. Our \href{https://youtu.be/pzT2fMwz428?t=321}{\VIDEO{accompanying video \cite{immeshvideo} (starting at 05:21)}} that visualizes the online mesh reconstruction process with sequence Kitti\_00 is available on YouTube.}
\label{table_four_dataset}
\footnotesize
\setlength\tabcolsep{4.8pt}
\renewcommand{\arraystretch}{1.00} 
\begin{tabular}{ccccccccccc}
	\toprule
\textbf{Sequece}              & \textbf{\begin{tabular}[c]{@{}c@{}}Traveling \\ length (m)\end{tabular}} & \textbf{\begin{tabular}[c]{@{}c@{}}Durations\\ (s)\end{tabular}} & \textbf{\begin{tabular}[c]{@{}c@{}}LiDAR \\ frames\end{tabular}} & \textbf{\begin{tabular}[c]{@{}c@{}}Meshing \\ mean/Std (ms)\end{tabular}} & \textbf{\begin{tabular}[c]{@{}c@{}}Localization\\ mean/Std (ms)\end{tabular}} & \textbf{\begin{tabular}[c]{@{}c@{}}Number of \\ vertices (m)\end{tabular}} & \textbf{\begin{tabular}[c]{@{}c@{}}Number of\\  facets(m)\end{tabular}} & \textbf{Scenarios} \\ 	\toprule 
Kitti\_00                     & 3,724.2                                                                  & 456                                                              & 4,541                                                            & 32.1 / 12.0                                                               & 49.0 / 11.7                                                                   & 3.33                                                                    & 7.70                                                                 & Urban city         \\
Kitti\_01                     & 2,453.2                                                                  & 146                                                              & 1,101                                                            & 34.5 / 10.5                                                               & 51.1 / 18.5                                                                   & 2.03                                                                    & 4.05                                                                 & High way           \\
Kitti\_02                     & 5,058.9                                                                  & 509                                                              & 4,661                                                            & 33.5 / 7.0                                                                & 36.2 / 9.5                                                                    & 4.39                                                                    & 10.03                                                                & Residential        \\
Kitti\_03                     & 560.9                                                                    & 88                                                               & 801                                                              & 28 / 7.1                                                                  & 49.0 / 12.2                                                                   & 0.73                                                                      & 1.55                                                                 & Countryside; Road  \\
Kitti\_04                     & 393.6                                                                    & 27                                                               & 271                                                              & 30.1 / 9.4                                                                & 42.4 / 12.9                                                                   & 0.41                                                                      & 0.85                                                                   & Urban city; Road   \\
Kitti\_05                     & 2,205.6                                                                  & 303                                                              & 2,761                                                            & 29.6 / 8.2                                                                & 38.7 / 11.5                                                                   & 2.17                                                                    & 4.95                                                                 & Residential        \\
Kitti\_06                     & 1,232.9                                                                  & 123                                                              & 1,101                                                            & 23.1 / 5.6                                                                & 56.9 / 9.7                                                                    & 0.89                                                                      & 1.89                                                                 & Urban city         \\
Kitti\_07                     & 2,453.2                                                                  & 114                                                              & 1,101                                                            & 20.7 / 7.4                                                                & 31.3 / 8.6                                                                    & 0.76                                                                      & 1.71                                                                 & Urban city         \\
Kitti\_08                     & 3,222.8                                                                  & 441                                                              & 4,071                                                            & 32.4 / 7.8                                                                & 45.7 / 17.7                                                                   & 3.56                                                                    & 7.94                                                                 & Urban city         \\
Kitti\_09                     & 1,705.1                                                                  & 171                                                              & 1,591                                                            & 34.5 / 7.5                                                                & 43.1 / 19.2                                                                   & 1.83                                                                    & 4.12                                                                 & Countryside; Road  \\
Kitti\_10                     & 919.5                                                                    & 132                                                              & 1,201                                                            & 23.4 / 6.9                                                                & 30.9 / 11.9                                                                   & 0.94                                                                      & 2.10                                                                 & Residential        \\
\toprule 
NCLT 2012-01-15               & 7,499.8                                                                  & 6739                                                             & 66,889                                                           & 26.3 / 14.1                                                               & 21.3 / 9.8                                                                    & 9.66                                                                    & 26.61                                                                & Campus; Indoor     \\
NCLT 2012-04-29               & 3,183.1                                                                  & 2598                                                             & 25,819                                                           & 25.4 / 13.9                                                               & 19.1 / 5.4                                                                    & 4.82                                                                    & 13.43                                                                & Campus             \\
NCLT 2012-06-15               & 4,085.9                                                                  & 3310                                                             & 32,954                                                           & 24.5 / 14.4                                                               & 22.3 / 7.7                                                                    & 6.36                                                                    & 17.47                                                                & Campus             \\
NCLT 2013-01-10               & 1,132.3                                                                  & 1024                                                             & 10,212                                                           & 20.2 / 12.5                                                               & 19.3 / 6.5                                                                    & 2.02                                                                    & 5.50                                                                 & Campus             \\
NCLT 2013-04-05               & 4,523.6                                                                  & 4167                                                             & 41,651                                                           & 20.6 / 13.8                                                               & 26.8 / 11.7                                                                   & 9.58                                                                    & 23.98                                                                & Campus             \\
\toprule 
NTU VIRAL eee\_01             & 265.3                                                                    & 398                                                              & 3,987                                                            & 11.2 / 6.7                                                                & 14.5 / 3.4                                                                    & 0.60                                                                      & 1.38                                                                 & Aerial; Outdoor    \\
NTU VIRAL nya\_01             & 200.6                                                                    & 396                                                              & 3,949                                                            & 9.4 / 5.3                                                                 & 10.2 / 1.7                                                                    & 0.54                                                                      & 1.24                                                                 & Aerial; Indoor     \\
NTU VIRAL rtp\_01             & 449.6                                                                    & 482                                                              & 4,615                                                            & 12.1 / 8.5                                                                & 10.9 / 2.6                                                                    & 0.72                                                                      & 2.03                                                                 & Aerial; Outdoor    \\
NTU VIRAL sbs\_01             & 222.1                                                                    & 354                                                              & 3,542                                                            & 11.4 / 8.0                                                                & 17.2 / 3.2                                                                    & 0.47                                                                      & 1.15                                                                 & Aerial; Outdoor    \\
NTU VIRAL tnp\_01             & 319.4                                                                    & 583                                                              & 5,795                                                            & 6.3 / 3.7                                                                 & 8.8 / 1.2                                                                     & 0.16                                                                      & 0.41                                                                   & Aerial; Indoor     \\
\toprule 
R$^3$LIVE hku\_campus\_00     & 190.6                                                                    & 202                                                              & 2,022                                                            & 12.0 / 7.3                                                                & 11.5 / 3.2                                                                    & 0.58                                                                      & 1.24                                                                 & Campus             \\
R$^3$LIVE hku\_campus\_01     & 374.6                                                                    & 304                                                              & 3,043                                                            & 20.4 / 12.6                                                               & 17.2 / 6.9                                                                    & 1.32                                                                    & 2.86                                                                 & Campus             \\
R$^3$LIVE hku\_campus\_02     & 354.3                                                                    & 323                                                              & 3,236                                                            & 13.5 / 6.4                                                                & 11.9 / 2.8                                                                    & 0.87                                                                      & 1.91                                                                 & Campus             \\
R$^3$LIVE hku\_campus\_03     & 181.2                                                                    & 173                                                              & 1,737                                                            & 12.2 / 5.7                                                                & 11.3 / 2.9                                                                    & 0.55                                                                      & 1,13                                                                 & Campus             \\
R$^3$LIVE hku\_main\_building & 1,036.9                                                                  & 1170                                                             & 11,703                                                           & 16.9 / 14.3                                                               & 12.5 / 8.0                                                                    & 3.03                                                                    & 6.80                                                                 & Indoor; Outdoor    \\
R$^3$LIVE hku\_park\_00       & 247.3                                                                    & 228                                                              & 2,285                                                            & 30.1 / 15.9                                                               & 12.6 / 3.7                                                                    & 0.92                                                                      & 2.38                                                                 & Cluttered field      \\
R$^3$LIVE hku\_park\_01       & 401.8                                                                    & 351                                                              & 3,520                                                            & 31.5 / 12.2                                                               & 12.6 / 3.9                                                                    & 1.67                                                                    & 3.96                                                                 & Cluttered field      \\
R$^3$LIVE hkust\_campus\_00   & 1,317.2                                                                  & 1073                                                             & 10,732                                                           & 26.0 / 12.8                                                               & 18.0 / 7.6                                                                    & 4.92                                                                    & 11.25                                                                & Campus             \\
R$^3$LIVE hkust\_campus\_01   & 1,524.3                                                                  & 1162                                                             & 11,629                                                           & 27.1 / 13.9                                                               & 16.8 / 6.7                                                                    & 5.35                                                                    & 12.64                                                                & Campus             \\
R$^3$LIVE hkust\_campus\_02   & 2,112.2                                                                  & 1618                                                             & 4,787                                                            & 26.7 / 14.5                                                               & 20.3 / 6.1                                                                    & 1.99                                                                   & 4.65                                                                 & Campus             \\
R$^3$LIVE hkust\_campus\_03   & 503.8                                                                    & 478                                                              & 16,181                                                           & 33.6 / 13.3                                                               & 21.0 / 5.3                                                                     & 7.67                                                                   & 18.25                                                                & Campus            
                                                     
	\\	\toprule                                                        
\end{tabular}
    \vspace{-0.5cm}
\end{table*}

\subsubsection{Result and analysis}
As presented in the video, benefiting from the accurate uncertainty models of the LiDAR point and plane that account for both LiDAR measurement noise and sensor pose estimation errors in our \textit{{localization}} module, ImMesh is able to provide the $6$ DoF pose estimation of high accuracy in real-time. Without any additional processing (i.e., loop detection), all of these two trials can close the loop itself after traveling \SI{957}{\meter} and \SI{391}{\meter}, respectively. In addition, with the efficient architecture design and careful engineering implementation on our {\textit{meshing}} module, the triangle mesh of the surrounding environment is incrementally reconstructed on the fly. With the live preview of real-time meshing, it informs users whether the data collection is sufficient enough for any part of the scene. This important function could lower the revisit chances and facilitate the collection process. 
Immediately after the data collection, the dense accurate triangle mesh of this scene would be available for analysis. Due to this reason, our system is named as the \textbf{Im}mediately \textbf{Mesh}ing (ImMesh).

\setlength{\textfloatsep}{-0.1cm}
\iftrue
\begin{table}[t]    
    \centering
    \captionsetup{justification=centering}
    \footnotesize
    \setlength\tabcolsep{3.0pt}
    \renewcommand{\arraystretch}{1.0} 
    \caption{Two ImMesh configurations for two types of LiDARs (i.e., mechanical and solid-state LiDAR).}
    \label{table_two_configurations}
    \begin{tabular}{ccccc}
        \toprule
        & \textbf{ Minimum point} & \textbf{ Size of region } & \textbf{ Size of voxel }  \\ 
        & \textbf{ distance $\xi$ (\si{\meter})} & \textbf{  $S_{\mathbf{R}}$ (\si{\meter})} & \textbf{ $S_{\mathbf{O}}$ (\si{\meter})} & \\ \toprule 
        \textbf{Mechanical LiDAR}         & 0.15      & 15.0       & 0.60               \\
        \textbf{Solid-state LiDAR}         & 0.10      & 10.0       & 0.40               \\ \toprule
    \end{tabular}
\end{table}
    \vspace{-0.2cm}
\begin{table}[t]    
    \centering
    \captionsetup{justification=centering}
    \footnotesize
    \setlength\tabcolsep{3.0pt}
    \renewcommand{\arraystretch}{0.95} 
    \caption{The average/maximum time of \textit{meshing} and \textit{localization} module for processing each LiDAR scan in four datasets.}
    \label{table_avg_max_time_dataset}
    \begin{tabular}{ccccc}
        \toprule
        & \textbf{Kitti} & \textbf{NCLT} & \textbf{NTU VIRAL} & \textbf{R$^3$LIVE} \\ 
        & \textbf{mean/max} & \textbf{mean/max} & \textbf{mean/max} & \textbf{mean/max} \\ \toprule
        \textbf{Meshing (ms)}         & 31.3 / 34.5      & 24.2 / 25.4       & 9.8 / 17.2            & 25.3 / 33.6         \\
        \textbf{Localization (ms)} & 42.2 / 56.9      & 22.3 / 26.8      & 11.9 / 17.2           & 16.6 / 21.0       \\ \toprule
    \end{tabular}
    \vspace{0.3cm}
\end{table}
\fi

\begin{figure*}[t]
    \centering
    \begin{minipage}{1.02\linewidth}
    \setlength{\belowcaptionskip}{-0.6cm}
    \includegraphics[width=1.00\linewidth]{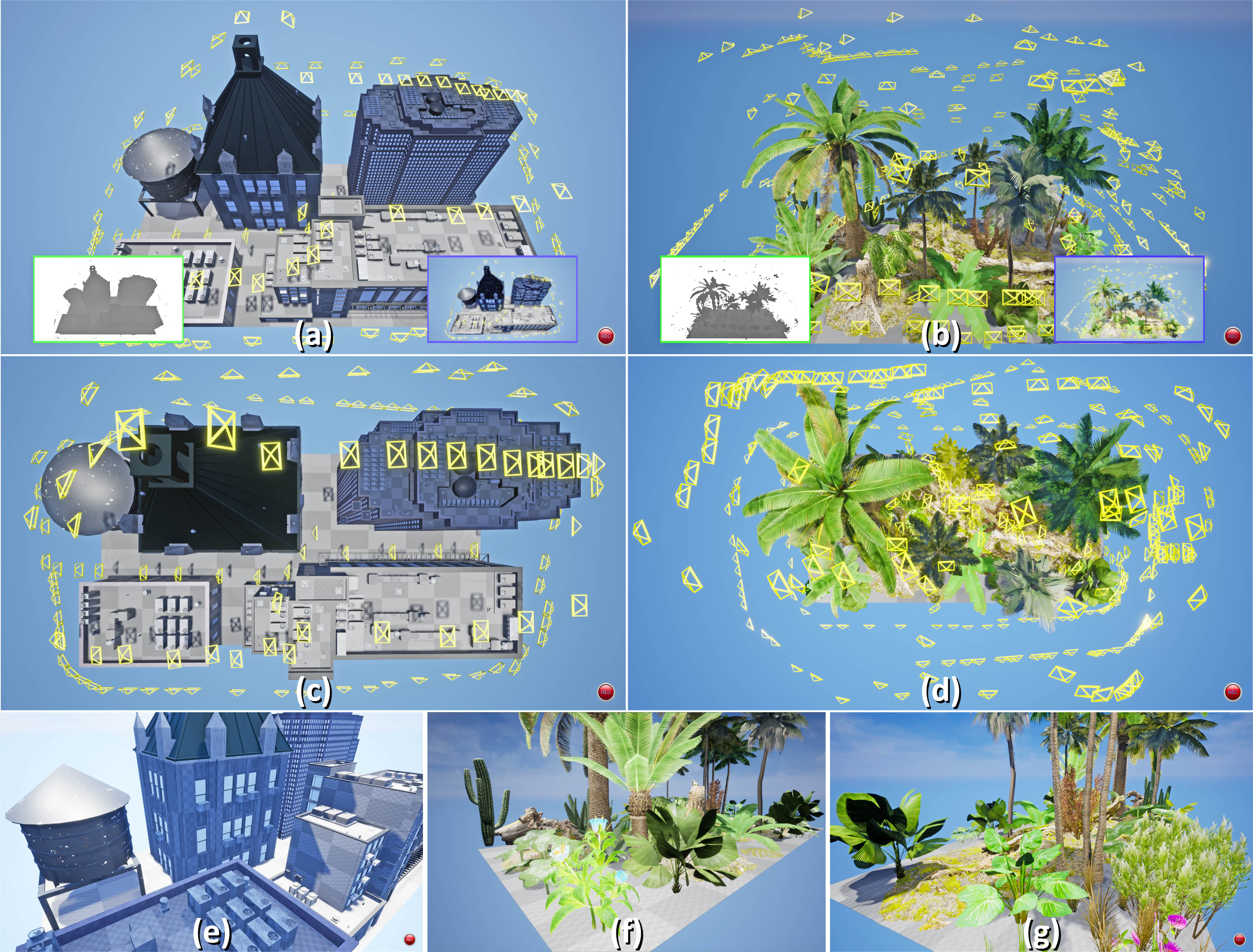}
    \caption{The screenshots of our \textit{Microsoft AirSim} simulator used for generating synthetic data.  (a, c, and e) show the ``Urban city" environments, while (b, d, f, and g) depict the ``Cluttered field" environments. The yellow frustums in (a$\sim$d) represent the poses of LiDAR sensor used to capture synthetic data. These frustums are set as invisible during data generation, as shown in (e-g). The images within the green and blue boxes in (a and d) respectively show the produced depth and RGB images.}
    \label{fig_airsim_simulator}
    \end{minipage}
    \vspace{0.3cm}
\end{figure*}

\subsection{Experiment-2: Extensive evaluation of ImMesh on public datasets with various types of LiDAR in different scenes}

With all the modules delicately designed for efficiency, both the \textit{localization} and \textit{meshing} modules easily achieve real-time performances on a standard multi-core CPU. In this experiment, we evaluate the average time consumption on four public datasets with the computation platform listed in Section \ref{sect_experimetn_setup}.

The four datasets we chose are: Kitti dataset\cite{geiger2012we}, NCTL dataset\cite{carlevaris2016university}, NTU VIRAL dataset\cite{nguyen2022ntu} and R$^3$LIVE dataset\cite{r3live}. They are collected in different scenarios ranging from structured urban buildings to field-cluttered complex environments (see TABLE \ref{table_four_dataset}), using various types of LiDARs that include mechanical spinning LiDAR of different channels and solid-state LiDAR of small FoV (see TABLE \ref{table_lidar_spec}). 
\subsubsection{Experiment setup}
ImMesh is robust to its parameter values, which requires minimal user-adjustable parameters to achieve good results without extensive parameter tuning. We benchmark ImMesh in four datasets with only two sets of configurations. The two configurations are reasonably required for adapting two classes of LiDARs (i.e., mechanical and solid-state LiDAR), as shown in TABLE~\ref{table_two_configurations}. Since the 3D points sampled by a solid-state LiDAR are distributed in a small sensor FoV, the accumulated point cloud of solid-state LiDAR usually has a higher density. Therefore, we set the minimum point distance and voxel size for solid-state LiDAR $1.5$ times smaller than those for mechanical LiDAR, as shown in TABLE~\ref{table_two_configurations}. We maintained the same configuration for the other setups except for some necessary adjustments to match the hardware setup.

\subsubsection{Result and analysis}
TABLE \ref{table_four_dataset} shows the detailed information (e.g., length, duration, scene) of each sequence, the average time consumption of our {\textit{localization}} and \textit{{meshing}} module in processing a LiDAR scan, and the number of vertices and facets of each reconstructed mesh. From Table \ref{table_four_dataset}, it is seen that the average cost-time of both \textit{localization} and \textit{meshing} modules are closely related to the density of the input LiDAR scan. To be detailed, the LiDAR of a higher channel has a much higher point sampling rate (see Table~\ref{table_lidar_spec}) which causes more data to be processed in each update of a LiDAR frame (e.g., more points in a voxel and more voxels activated in each frame). Besides, the processing time varies among different scenarios for the same set of datasets. The sequences sampled in a high-way or field environment (e.g., Kitti\_01, Kitti\_09) usually have a longer LiDAR sampling range, leading to more points per frame to be processed. Thanks to the efficient data structures (e.g., ikd-Tree, hash tables) and parallelism strategy, which allows us to perform the state estimation and incremental mesh reconstruction simultaneously, the time consumption of large-scale datasets is bounded in an acceptable value ($\leq$\SI{35}{\milli\second} for meshing, $\leq$\SI{49}{\milli\second} for localization). 

The average and maximum time consumption of ImMesh in the four datasets are shown in TABLE \ref{table_avg_max_time_dataset}, reflecting that our system satisfies the real-time requirement even with different types of LiDARs and scenarios. Notice that the LiDAR frame rate are \SI{10}{\hertz} for all datasets, and our \textit{meshing} and \textit{{localization}} modules run in parallel (see Section \ref{sect_parallelism}). 

\begin{figure*}[h]
\centering
\begin{minipage}{0.93\linewidth}
    \centering
   \includegraphics[width=0.96\linewidth]{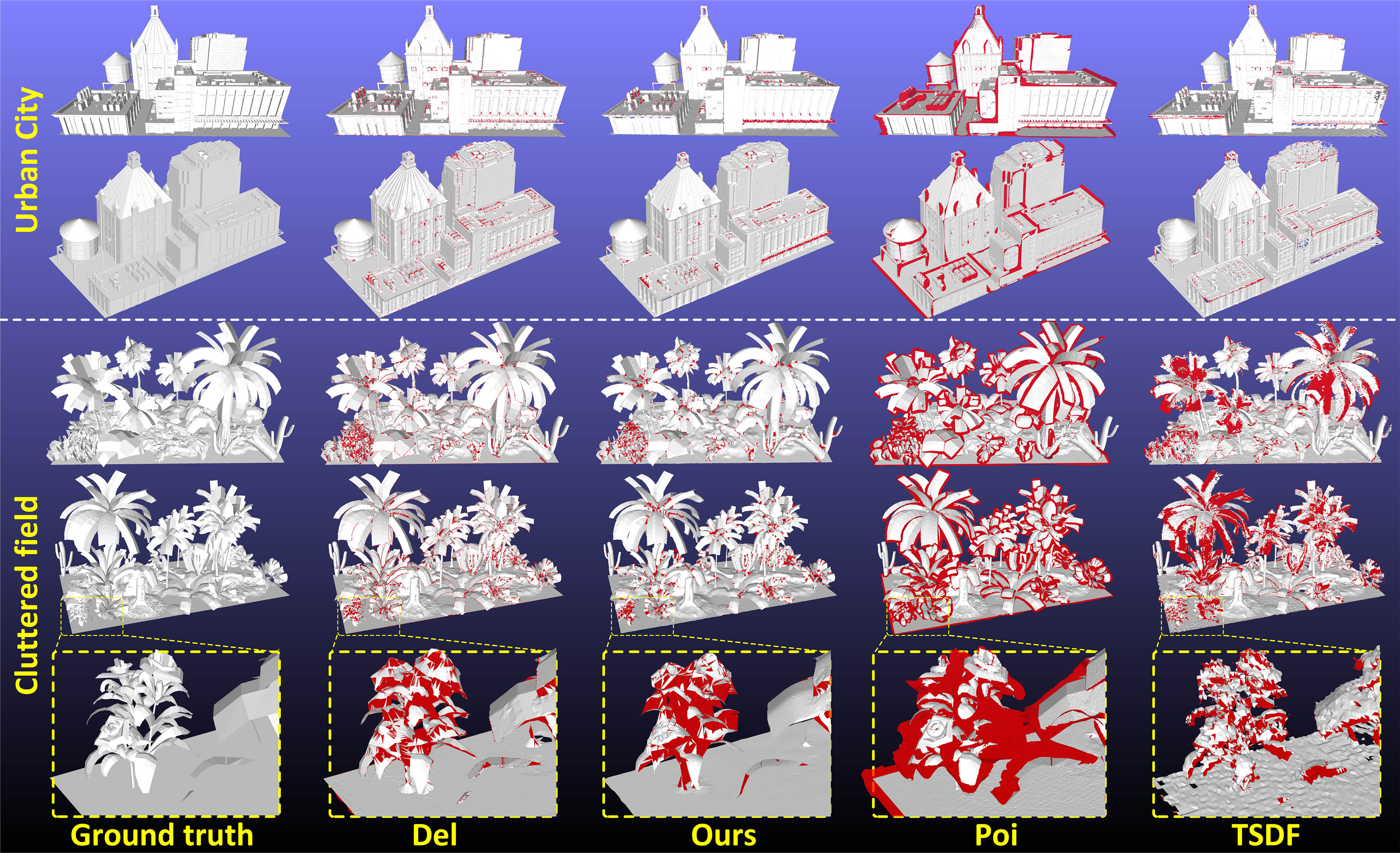}
       \caption{The qualitative comparison of ground truth and four evaluated methods, which are tested with the depth images resolution of $640\times 480$. \HL{The facets colored in red represent surfaces that have been incorrectly reconstructed, with $80\%$ of their sampling points not lying on the ground truth surface (i.e., the distances between these points and the nearest ground truth surface are larger than \SI{5}{\centi\meter}).}}
        \label{fig_simulator_comp}
    \centering
    \vspace{0.05cm}
    \includegraphics[width=0.96\linewidth]{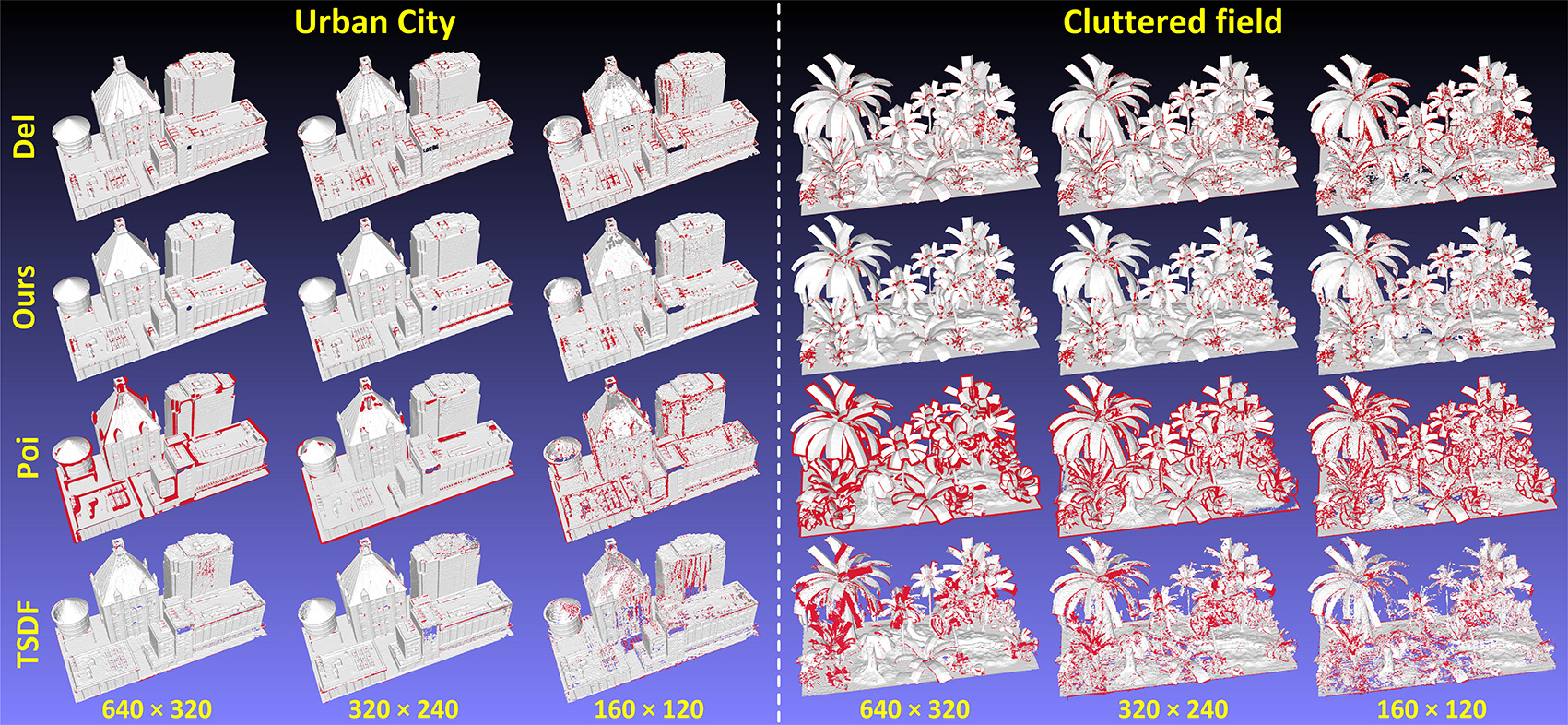}
    \caption{The qualitative comparisons of four methods that evaluated with depth images of different resolutions. \HL{The facets colored in red represent surfaces that have been incorrectly reconstructed, with $80\%$ of their sampling points not lying on the ground truth surface.}}
     \label{fig_simulator_mul}
    \vspace{-0.6cm}
     \end{minipage}
\end{figure*}

\begin{table*}[h]
    \centering
    \caption{The quantitative evaluation result with real-world data from \textit{Complex Urban Dataset}. The $\uparrow$ denotes larger is better while $\downarrow$ indicates lower is better.}
    \label{table_gt_urban_res}
\footnotesize
\setlength\tabcolsep{2.8pt}
\renewcommand{\arraystretch}{1.00} 
\begin{tabular}{cccccccccccc}
\hline
\multirow{3}{*}{\textbf{Sequence}} & \multirow{3}{*}{\textbf{\begin{tabular}[c]{@{}c@{}}Traveling\\      length (Km)\end{tabular}}} & \multirow{3}{*}{\textbf{\begin{tabular}[c]{@{}c@{}}Number of \\      LiDAR frames\end{tabular}}} & \multirow{3}{*}{\textbf{Method}} & \multirow{3}{*}{\textbf{\begin{tabular}[c]{@{}c@{}}Cost time ↓ \\       (hour:min:sec)\end{tabular}}} & \multicolumn{2}{|c}{\textbf{Fairness}}                                                          & \multicolumn{5}{|c}{\textbf{Correctness}}                                                                                                                                                                                                                                                                                   \\ \cline{6-12} 
                                   &                                                                                                &                                                                                                  &                                  &                                                                                                       & \multicolumn{1}{|c}{\textbf{\begin{tabular}[c]{@{}c@{}}Max-Min\\       angle (°) ↓\end{tabular}}} & \textbf{C2SE ↓} & \multicolumn{1}{|c}{\textbf{\begin{tabular}[c]{@{}c@{}}Compleness\\      (m) ↓\end{tabular}}} & \textbf{\begin{tabular}[c]{@{}c@{}}Accuracy\\      (m) ↓\end{tabular}} & \textbf{\begin{tabular}[c]{@{}c@{}}Recall\\      (\%) ↑\end{tabular}} & \textbf{\begin{tabular}[c]{@{}c@{}}Precision \\      (\%) ↑\end{tabular}} & \textbf{F-score ↑} \\
\toprule
\multirow{2}{*}{Urban01}           & \multirow{2}{*}{11.72}                                                                         & \multirow{2}{*}{13846}                                                                           & Poi                              & 09:58:30                                                                                              & {60.1014}                                                             & 0.9760          & 0.0632                                                                   & 0.0724                                                                 & 0.8554                                                                & 0.7563                                                                    & 0.8028             \\
                                   &                                                                                                &                                                                                                  & ImMesh (ours)                     & \textbf{00:05:39}                                                                                     & \textbf{56.2941}                                                             & \textbf{0.8630} & \textbf{0.0404}                                                          & \textbf{0.0568}                                                        & \textbf{0.9477}                                                       & \textbf{0.8260}                                                           & \textbf{0.8827}    \\
\toprule
\multirow{2}{*}{Urban02}           & \multirow{2}{*}{4.20}                                                                          & \multirow{2}{*}{8961}                                                                            & Poi                              & 05:49:36                                                                                              & {59.9695}                                                             & 0.9739          & 0.0792                                                                   & 0.0822                                                                 & 0.8818                                                                & 0.7261                                                                    & 0.7964             \\
                                   &                                                                                                &                                                                                                  & ImMesh (ours)                     & \textbf{00:03:01}                                                                                     & \textbf{57.3564}                                                             & \textbf{0.8605} & \textbf{0.0392}                                                          & \textbf{0.0556}                                                        & \textbf{0.9623}                                                       & \textbf{0.8398}                                                           & \textbf{0.8968}    \\
\toprule
\multirow{2}{*}{Urban03}           & \multirow{2}{*}{3.06}                                                                          & \multirow{2}{*}{9091}                                                                            & Poi                              & 04:55:04                                                                                              & {60.1614}                                                             & 0.9770          & 0.0070                                                                   & 0.0059                                                                 & 0.8871                                                                & 0.7754                                                                    & 0.8275             \\
                                   &                                                                                                &                                                                                                  & ImMesh (ours)                     & \textbf{00:03:07}                                                                                     & \textbf{57.4131}                                                             & \textbf{0.8628} & \textbf{0.0398}                                                          & \textbf{0.0564}                                                        & \textbf{0.9597}                                                       & \textbf{0.8359}                                                           & \textbf{0.8935}   
\\\toprule
\end{tabular}
     \vspace{0.1cm}
    \caption{The quantitative evaluation result with synthetic data generated with \textit{Microsoft AirSim} simulator.}
    \label{table_gt_airsim_res}
\footnotesize
\setlength\tabcolsep{4.8pt}
\renewcommand{\arraystretch}{1.05} 
\begin{tabular}{ccccccccccc}
\hline
\multirow{3}{*}{\textbf{Method}} & \multirow{3}{*}{\textbf{Scenario}} & \multirow{3}{*}{\textbf{Resolution}} & \multirow{3}{*}{\textbf{\begin{tabular}[c]{@{}c@{}}Cost time \\      (min:sec) ↓\end{tabular}}} & \multicolumn{2}{|c|}{\textbf{Fairness}}                                                          & \multicolumn{5}{c}{\textbf{Correctness}}                                                                                                                                                                                                                                                                                   \\ \cline{5-11} 
                                 &                                    &                                      &                                                                                                 & \multicolumn{1}{|c}{\textbf{\begin{tabular}[c]{@{}c@{}}Max-Min\\       angle (°) ↓\end{tabular}}} & \multicolumn{1}{c|}{\textbf{C2SE ↓}}  & \textbf{\begin{tabular}[c]{@{}c@{}}Compleness\\      (m) ↓\end{tabular}} & \textbf{\begin{tabular}[c]{@{}c@{}}Accuracy\\      (m) ↓\end{tabular}} & \textbf{\begin{tabular}[c]{@{}c@{}}Recall\\      (\%) ↑\end{tabular}} & \textbf{\begin{tabular}[c]{@{}c@{}}Precision \\      (\%) ↑\end{tabular}} & \textbf{F-score ↑} \\ 
\hline
Del                              & Urban city                         & 640 × 480                            & 17:51                                                                                           & \textbf{48.5148}                                                             & \textbf{0.7825} & \textbf{0.0883}                                                          & 0.0341                                                                 & \textbf{0.7976}                                                       & 0.7976                                                                    & \textbf{0.7976}    \\
ImMesh (ours)                    & Urban city                         & 640 × 480                            & 00:31                                                                                           & 48.0909                                                                      & 0.7843          & 0.1002                                                                   & \textbf{0.0265}                                                        & 0.7290                                                                & \textbf{0.8525}                                                           & 0.7859             \\
Poi                              & Urban city                         & 640 × 480                            & 15:42                                                                                           & 53.4110                                                                      & 0.8664          & 0.1094                                                                   & 0.2244                                                                 & 0.7367                                                                & 0.7798                                                                    & 0.7576             \\
TSDF                             & Urban city                         & 640 × 480                            & \textbf{00:25}                                                                                  & 64.7606                                                                      & 1.0530          & 0.1506                                                                   & 0.0361                                                                 & 0.5859                                                                & 0.8665                                                                    & 0.6991             \\
\hline
Del                              & Urban city                         & 320 × 240                            & 07:22                                                                                           & \textbf{49.1746}                                                             & \textbf{0.7960} & \textbf{0.0928}                                                          & 0.0515                                                                 & \textbf{0.7685}                                                       & 0.7470                                                                    & \textbf{0.7576}    \\
ImMesh (ours)                    & Urban city                         & 320 × 240                            & \textbf{00:23}                                                                                  & 52.9000                                                                      & 0.8235          & 0.1002                                                                   & \textbf{0.0265}                                                        & 0.7290                                                                & \textbf{0.7821}                                                           & 0.7546             \\
Poi                              & Urban city                         & 320 × 240                            & 04:42                                                                                           & 52.7233                                                                      & 0.8566          & 0.1216                                                                   & 0.0788                                                                 & 0.6845                                                                & 0.6904                                                                    & 0.6875             \\
TSDF                             & Urban city                         & 320 × 240                            & 00:25                                                                                           & 64.4962                                                                      & 1.0474          & 0.1544                                                                   & 0.0655                                                                 & 0.4994                                                                & 0.7397                                                                    & 0.5962             \\
\hline
Del                              & Urban city                         & 160 × 120                            & 02:02                                                                                           & \textbf{49.8635}                                                             & \textbf{0.8098} & \textbf{0.1186}                                                          & 0.0914                                                                 & \textbf{0.6822}                                                       & 0.5574                                                                    & \textbf{0.6135}    \\
ImMesh (ours)                     & Urban city                         & 160 × 120                            & \textbf{00:19}                                                                                  & 54.4587                                                                      & 0.8466          & 0.1341                                                                   & \textbf{0.0834}                                                        & 0.5493                                                                & 0.5914                                                                    & 0.5696             \\
Poi                              & Urban city                         & 160 × 120                            & 01:03                                                                                           & 54.6500                                                                      & 0.9052          & 0.1849                                                                   & 0.1453                                                                 & 0.4777                                                                & \textbf{0.6159}                                                           & 0.5381             \\
TSDF                             & Urban city                         & 160 × 120                            & 00:24                                                                                           & 65.1098                                                                      & 1.0564          & 0.2802                                                                   & 0.2508                                                                 & 0.3352                                                                & 0.4799                                                                    & 0.3947             \\
\hline
Del                              & Cluttered field                    & 640 × 480                            & 21:14                                                                                           & \textbf{56.6578}                                                             & \textbf{0.8304} & \textbf{0.2767}                                                          & \textbf{0.0496}                                                        & \textbf{0.7489}                                                       & 0.7036                                                                    & \textbf{0.7255}    \\
ImMesh (ours)                    & Cluttered field                    & 640 × 480                            & 00:33                                                                                           & 57.1687                                                                      & 0.8558          & 0.2953                                                                   & 0.0519                                                                 & 0.7027                                                                & \textbf{0.7404}                                                           & 0.7211             \\
Poi                              & Cluttered field                    & 640 × 480                            & 24:31                                                                                           & 59.5750                                                                      & 0.9649          & 0.3052                                                                   & 0.3960                                                                 & 0.6981                                                                & 0.7009                                                                    & 0.6995             \\
TSDF                             & Cluttered field                    & 640 × 480                            & \textbf{00:24}                                                                                  & 65.4224                                                                      & 1.0882          & 0.4130                                                                   & 0.4270                                                                 & 0.4837                                                                & 0.4936                                                                    & 0.4886             \\
\hline
Del                              & Cluttered field                    & 320 × 240                            & 07:38                                                                                           & \textbf{57.9700}                                                                      & \textbf{0.8526} & \textbf{0.2919}                                                          & 0.0882                                                                 & 0.5416                                                                & \textbf{0.7198}                                                           & 0.6181             \\
ImMesh (ours)                    & Cluttered field                    & 320 × 240                            & 00:25                                                                                           & 57.6159                                                                      & 0.8603          & 0.3105                                                                   & \textbf{0.0784}                                                        & \textbf{0.6146}                                                       & 0.6404                                                                    & \textbf{0.6272}    \\
Poi                              & Cluttered field                    & 320 × 240                            & 10:28                                                                                           & 59.4470                                                             & 0.9722          & 0.3620                                                                   & 0.3395                                                                 & 0.5630                                                                & 0.5506                                                                    & 0.5567             \\
TSDF                             & Cluttered field                    & 320 × 240                            & \textbf{00:23}                                                                                  & 65.7206                                                                      & 1.0892          & 0.5268                                                                   & 0.4784                                                                 & 0.2114                                                                & 0.2567                                                                    & 0.2319             \\
\hline
Del                              & Cluttered field                    & 160 × 120                            & 01:56                                                                                           & 59.4879                                                                      & \textbf{0.8785} & \textbf{0.3438}                                                          & 0.1781                                                                 & 0.3947                                                                & \textbf{0.5696}                                                           & \textbf{0.4663}    \\
ImMesh (ours)                    & Cluttered field                    & 160 × 120                            & \textbf{00:21}                                                                                  & 60.1208                                                                      & 0.8970          & 0.3512                                                                   & \textbf{0.1694}                                                        & \textbf{0.4200}                                                       & 0.4863                                                                    & 0.4507             \\
Poi                              & Cluttered field                    & 160 × 120                            & 01:07                                                                                           & \textbf{59.1544}                                                             & 0.9744          & 0.3541                                                                   & 0.4164                                                                 & 0.3775                                                                & 0.3820                                                                    & 0.3797             \\
TSDF                             & Cluttered field                    & 160 × 120                            & 00:23                                                                                           & 65.1832                                                                      & 1.0815          & 0.5561                                                                   & 0.3681                                                                 & 0.2099                                                                & 0.2944                                                                    & 0.2451            
\\ \toprule
\end{tabular}
    \vspace{-0.5cm}
\end{table*}

\begin{figure*}[t]
    \centering
    \setlength{\belowcaptionskip}{-0.5cm}
    \includegraphics[width=1.0\linewidth]{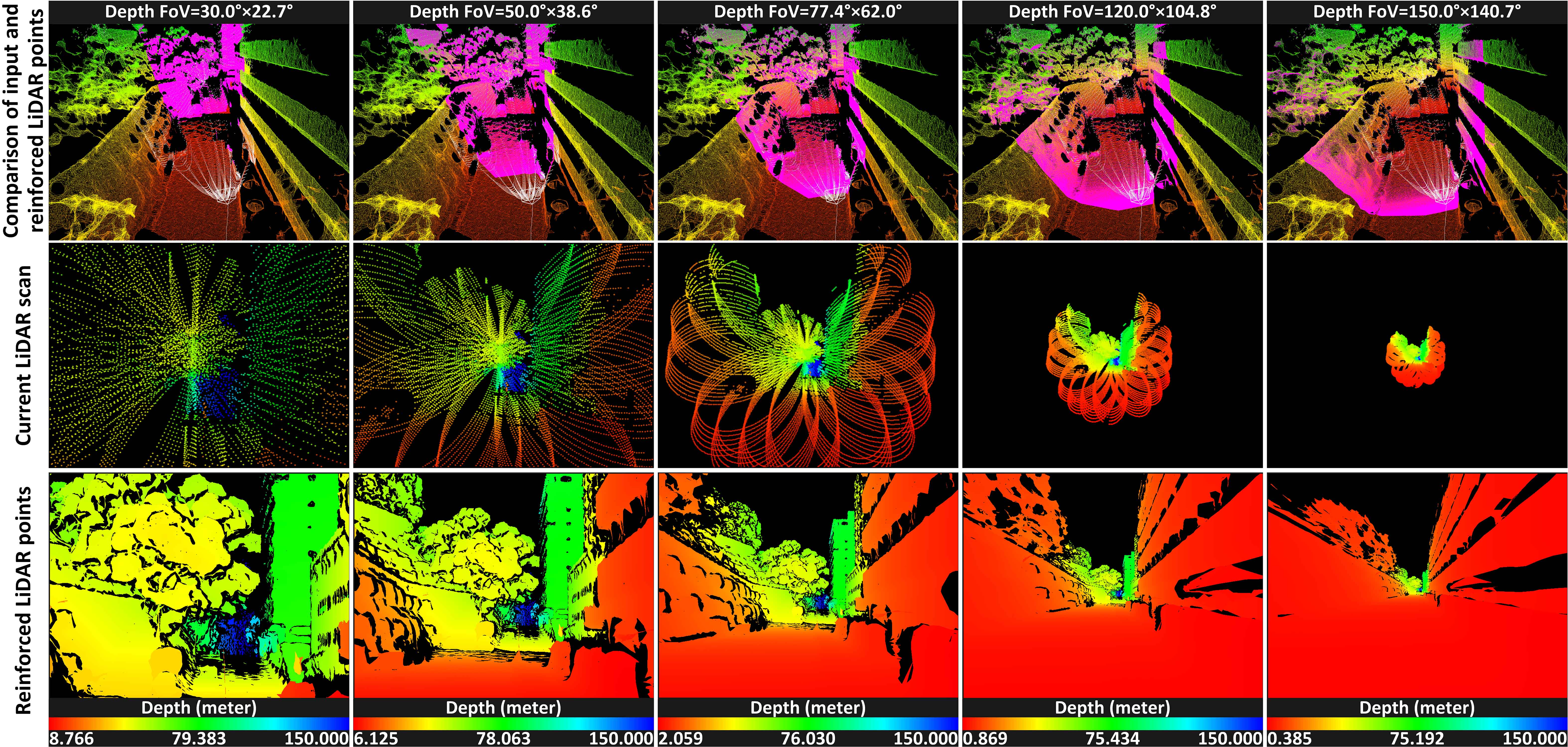}
    \caption{ The first row of images shows the comparisons between a raw LiDAR frame (colored in white) and our reinforced points (colored in magenta) under different sets of rasterizing FoV. The second and third rows of images show the comparisons of raw and reinforced points after projection on the current sensor frame. For more detailed visualizations of this process, please refer to our \href{https://youtu.be/pzT2fMwz428?t=499}{\VIDEO{accompanying video \cite{immeshvideo} (starting at 08:19)}} on YouTube.}
    \label{fig_LiDAR_point_cloud_reinforcement}
\end{figure*}

\subsection{Experiment-3: Quantitative evaluation of ImMesh}\label{section_exp_3}

In this experiment, we use both real-world and synthetic data to conduct the quantitative evaluations of ImMesh, by comparing it against existing reconstruction methods.


\subsubsection{Preparation of large-scale, real-world data}\label{sect_prepare_realword_data}

We conducted a quantitative evaluation using large-scale real-world LiDAR data collected from the \textit{Complex Urban Dataset} \cite{jeong2019complex}. This dataset provides a high-quality set of ground truth LiDAR poses and ground truth point clouds, which enables a comprehensive assessment of our proposed method and existing baselines.  The detailed traveling length and the number of LiDAR frames of tested sequences are shown in TABLE \ref{table_gt_urban_res}.

\subsubsection{Preparation of synthetic data}\label{sect_prepare_simulated_data}
To further evaluate the performance of all the methods under diverse scenarios, with varying levels of clutteredness, we generated synthetic data using the \textit{Microsoft AirSim} simulator\cite{shah2018airsim}. The screenshots of our simulating scenarios are presented in Fig. \ref{fig_airsim_simulator}, where we prepared two typical  environments: ``Urban city" (Fig. \ref{fig_airsim_simulator}(a, c, and e)) and ``Cluttered field" (Fig.~\ref{fig_airsim_simulator}(b, d, f, and g)), both of which have dimensions of \SI{20}{\meter}$\times$\SI{10}{\meter}$\times$\SI{8}{\meter}. The ``Urban city" environment consists of structured objects, such as buildings, towers, and water tanks, providing a realistic representation of an urban setting. On the other hand, the ``Cluttered field" environment incorporates a diverse range of plants, including trees, flowers, grasses, and other vegetation, creating a more complex and cluttered scenario.

To simulate point clouds collected by a real LiDAR, we unproject the 3D points from the depth image. The depth images are obtained by querying the \textit{AirSim}'s API, specifically the images shown within the green box in Fig. \ref{fig_airsim_simulator}(a and b). The depth image has a field of view (FoV) of $120^\circ\times 80^\circ$. We manually positioned the poses, represented by the yellow frustums in Fig.~\ref{fig_airsim_simulator}(a $\sim$ d), to ensure that the generated point cloud covers most of the surfaces in the scene. Additionally, we simulate LiDAR data with different point cloud densities by generating data using three different sets of depth image resolutions: $640\times 480$, $320\times 240$, and $160\times 120$, as shown in TABLE \ref{table_gt_airsim_res}.

\subsubsection{{Experiment setup}}
In this experiment, we performed a comprehensive evaluation of meshing ability among our work and existing mesh reconstruction baselines, which includes a TSDF-based method implemented by \textit{Point cloud library (PCL)}\cite{rusu20113d} with GPU acceleration, Delaunay triangulation and graph cut based method implemented by \textit{OpenMVS}\cite{openmvs2020}, and the official implementation of  Poisson surface reconstruction\cite{kazhdan2006poisson, kazhdan2013screened}.


We conducted the evaluation of these methods on a desktop PC equipped with an \textit{Intel i7-9700K} CPU, 64Gb RAM, and an \textit{Nvidia 2080 Ti} GPU with 12Gb of graphics memory. We fed online reconstruction method \textit{ImMesh} and TSDF-based (\textit{TSDF}) methods with LiDAR points frame by frame. To mitigate the impact of pose estimation errors on meshing results, we disabled the pose estimation module and provided the ground truth poses to the online mesh reconstruction methods \textit{ImMesh} and \textit{TSDF}. For the offline mesh reconstruction methods, namely Delaunay triangulation (\textit{Del}) and Poisson surface reconstruction (\textit{Poi}), we fed them with the accumulated point cloud from all frames. Additionally, to address the issue of uneven point cloud density, which can result in errors when calculating normals for \textit{Poi}, and to prevent \textit{Del} from reconstructing small facets that could bias accuracy calculations. We leverage a voxel grid filter with a leaf size of $\SI{1.0}{\centi\meter}\times\SI{1.0}{\centi\meter}\times\SI{1.0}{\centi\meter}$ to downsample the accumulated point cloud before providing it as input to both \textit{Poi} and \textit{Del}.

Due to the limited graphics memory (12Gb for \textit{Nvidia 2080 Ti}), we set the \textit{TSDF} cell size as \SI{0.2}{\meter} such that \textit{TSDF} can utilize the GPU acceleration while preserving satisfying precision in the mesh reconstruction. For our ImMesh, the parameter configuration for solid-state LiDAR is used, as shown in TABLE~\ref{table_two_configurations}. For \textit{Poi}, we set the octree level as $12$ and removed large hulls by deleting facets with one of their edges longer than \SI{15.0}{\centi\meter}. For other configurations of all methods, we set them as their default configuration. It is noted that other than \textit{TSDF} using GPUs for acceleration, the rest methods, \textit{Del}, \textit{Poi}, and ours, use the CPU only. We compare the efficiency of four methods by evaluating their time consumption in reconstructing the mesh. For online methods (i.e., \textit{TSDF} and ours), we accumulate the processing time of all frames, while for offline methods (i.e., \textit{Poi} and \textit{Del}), we count the total time in processing the offline data. The results of their time consumption are listed in TABLE~\ref{table_gt_urban_res} and TABLE~\ref{table_gt_airsim_res}.

\subsubsection{Evaluation of fairness}\label{sect_exp3_res_ana_fairness}
In this experiment, we employ the triangle fairness criteria to evaluate the quality of reconstructed triangle facets. This evaluation involves analyzing the average error of the maximum and minimum interior angles of the triangles (as utilized in work \cite{lawson1977software}), which we refer to as the \textit{Max-Min angle} in TABLE \ref{table_gt_urban_res} and TABLE \ref{table_gt_airsim_res}. Additionally, we consider the average ratio of the circumradius to the shortest edge length (referred to as \textit{C2SE} in Tables) as used in works \cite{shewchuk1997delaunay, li2003generating}. A lower value for both the \textit{Max-Min angle} and \textit{C2SE} indicates higher mesh quality, as it signifies that the triangle facets are closer to being equilateral.

In the evaluation with large-scale, real-world data, the results for \textit{Del} and \textit{TSDF} methods were not available due to specific limitations: 1) For \textit{Del}, we encountered difficulties when running it with the \textit{Complex Urban Dataset}. Despite multiple attempts, the \textit{Del} method either crashed midway or failed to produce any result after running for over three days. 2) As for \textit{TSDF}, allocating the voxels requires a massive amount of graphics memory. This exceeds the capabilities of our hardware platforms, particularly for sequences in Table~\ref{table_gt_urban_res} with a traveling length of over 3 kilometers.

As indicated by the fairness metrics listed in TABLE \ref{table_gt_urban_res} and \ref{table_gt_airsim_res}, we can conclude that leveraging Delaunay triangulation eliminates the formation of sliver triangles. The \textit{Del} method demonstrates the best results in this regard. Following that is \textit{ImMesh}, which utilizes Delaunay triangulation for meshing the point set after dimension reduction through projection. On the other hand, the meshes reconstructed by the \textit{Poi} and \textit{TSDF} methods, which employ the marching cubes algorithm, exhibit inferior results. This is due to the inherent limitation of the marching cubes algorithm  \cite{lorensen1987marching}, which generates sliver triangles when a facet is positioned closely and nearly parallel to the edges of the cube.

\subsubsection{Evaluation of correctness}\label{sect_exp3_res_ana_accuracy}
For the quantitative evaluation of the methods' correctness in reconstructing the mesh, we utilized 3D geometry metrics as employed in works NeuralRecon \cite{sun2021neuralrecon} and Atlas \cite{murez2020atlas}. These metrics encompass the following measurements: \textit{accuracy}, \textit{completeness}, \textit{precision}, \textit{recall}, and \textit{F-score}. The calculations for these metrics are as follows:
\begin{align}
    \hspace{-0.4cm}\text{\textit{Accuracy}:} & ~~ \mathtt{mean}_{\mathbf{p}\in \boldsymbol{\mathcal{P}}}\large( \mathtt{min}_{\mathbf{p}^{*}\in \boldsymbol{\mathcal{P}}^{*}} || \mathbf{p} - \mathbf{p}^{*}  || \large) \nonumber \\
    \hspace{-0.4cm}\text{\textit{Completeness}:} &  ~~  \mathtt{mean}_{\mathbf{p}^{*}\in \boldsymbol{\mathcal{P}}^{*}}\large( \mathtt{min}_{\mathbf{p}\in \boldsymbol{\mathcal{P}}} || \mathbf{p} - \mathbf{p}^{*}  || \large) \nonumber \\
    \hspace{-0.4cm}\text{\textit{Precision}:} & ~~  \mathtt{mean}_{\mathbf{p}\in \boldsymbol{\mathcal{P}}}\large( \mathtt{min}_{\mathbf{p}^{*}\in \boldsymbol{\mathcal{P}}^{*}} || \mathbf{p} - \mathbf{p}^{*}  || < 0.05 \large)  \nonumber \\
    \hspace{-0.4cm}\text{\textit{Recall}:} & ~~ \mathtt{mean}_{\mathbf{p}^{*}\in \boldsymbol{\mathcal{P}}^{*}}\large( \mathtt{min}_{\mathbf{p}\in \boldsymbol{\mathcal{P}}} || \mathbf{p} - \mathbf{p}^{*}|| < 0.05  \large) \nonumber \\
    \text{\textit{F-score}:} & ~~ \dfrac{2\times \text{\textit{Precision}} \times \text{\textit{Recall}} }{\text{\textit{Precision}} + \text{\textit{Recall}}} \nonumber
\end{align}
where $\boldsymbol{\mathcal{P}}$ refers to the point cloud obtained by uniformly sampling the reconstructed mesh generated by the method under evaluation. This point cloud is sampled at a spatial resolution of \SI{0.01}{\meter}. On the other hand, $\boldsymbol{\mathcal{P}}^*$ represents the downsampled ground truth point cloud. It is also downsampled at a spatial resolution of \SI{0.01}{\meter}.


The quantitative evaluation results for metrics such as \textit{accuracy}, \textit{completeness}, \textit{precision}, \textit{recall}, and \textit{F-score} are provided in TABLE \ref{table_gt_airsim_res}.  We can observe that \textit{Del} achieves the highest overall correctness in constructing the mesh of the scene. Following that, \textit{ImMesh} demonstrates slightly lower precision. Then, \textit{Poi} exhibits even lower mesh correctness, and \textit{TSDF} shows the lowest correctness among all methods.

The qualitative comparison results of the four benchmarked methods evaluated with synthetic data are presented in Fig. \ref{fig_simulator_comp} and Fig. \ref{fig_simulator_mul}. In these figures, the red facets represent incorrectly reconstructed surfaces with $80\%$ of their sampling points not lying on a ground truth surface \HL{(i.e., the distances between these points and the nearest ground truth surface are larger than \SI{5}{\centi\meter})}. Among the evaluated methods, \textit{Del} and \textit{ImMesh} exhibit comparable results in reconstructing the mesh of scenes well. In contrast, \textit{Poi} exhibits lower mesh correctness due to the presence of unwanted facets at the sharp edges of the models, as indicated in the fourth column of Fig. \ref{fig_simulator_comp}. The \textit{TSDF} method shows the lowest results with the appearance of holes on the reconstructed surface, as observed in the roofs of buildings and the leaves of trees shown in the fifth column of Fig. \ref{fig_simulator_comp}.


When reconstructing complex and small objects in the scene, such as the flower in the ``Cluttered field" environment, as depicted in the RGB image shown in the bottom-left corner of Fig. \ref{fig_airsim_simulator}(f) and the corresponding mesh models displayed in the fifth row of Fig. \ref{fig_simulator_comp}. \textit{Del}, \textit{TSDF} and \textit{ImMesh} fail to recover the details of surface well. This limitation arises from different factors for each method: \textit{Del} requires a large number of camera-to-point correspondences to extract intricate surface details, which may pose challenges when dealing with complex and tiny objects. \textit{TSDF} and \textit{ImMesh} are constrained by the fixed voxel size, which can not reconstruct the details of surfaces whose size is smaller than voxel. What is worth mentioning is that we found \textit{Poi} can recover the details of the flower's petals well. This is achieved through the use of a scalable resolution based on an octree structure, which allows \textit{Poi} to adapt its resolution for reconstructing small and intricate surfaces.

In addition, as observed in Fig. \ref{fig_simulator_mul} and with the metrics listed in Table \ref{table_gt_airsim_res}, we can see that as the point cloud becomes sparser (due to lower resolution depth images), the correctness of the reconstruction methods decreases accordingly. However, both \textit{Del} and \textit{ImMesh} demonstrate stronger robustness in resiliently handling the drop in point cloud density. On the other hand, the meshes reconstructed by \textit{Poi} and \textit{TSDF} exhibit discontinuities and contain more holes and gaps when compared to the results of \textit{Del} and \textit{ImMesh}.
 
Lastly, in the evaluation with real-world data from \textit{Complex Urban Dataset}\cite{jeong2019complex}, we discovered that the mesh reconstructed by \textit{Poi} also exhibits unwanted facets appearing at the edges of objects such as buildings and trees. These undesirable facets, as indicated by the red facets in Fig. \ref{fig_simulator_comp} and Fig. \ref{fig_simulator_mul} for \textit{Poi}, have a negative impact on the overall correctness of the reconstruction. As a result, \textit{Poi} performs inferiorly across all evaluated correctness metrics when compared to \textit{ImMesh}, as shown in TABLE \ref{table_gt_urban_res}.

\subsubsection{Evaluation of runtime performance}\label{sect_exp3_res_ana_runtime}

According to the \textit{cost time} listed in TABLE~\ref{table_gt_urban_res}, it is clear that \textit{ImMesh} demonstrates a significant advantage in terms of runtime performance when evaluated with large-scale sequences. The execution time of \textit{ImMesh} is only $0.93\% \sim 1.06\%$ of that of \textit{Poi}.

TABLE \ref{table_gt_airsim_res} displays the average time consumption of the four benchmarked methods when evaluated with synthetic data. The online methods, \textit{ImMesh} and \textit{TSDF}, exhibit similar runtime performance. In contrast, the offline methods (\textit{Del} and \textit{Poi}) consume significantly more time, ranging from $5$ to $40$ times longer than the online methods (\textit{TSDF} and \textit{ImMesh}). Notably, \textit{TSDF} achieves comparable runtime performance to our method with the assistance of an \textit{Nvidia 2080 Ti} GPU, highlighting the high computational efficiency of our \textit{ImMesh} framework compared to the other three methods.

\subsubsection{Summary}

Based on the results and analysis regarding runtime performance, fairness, and correctness, we have reached the following conclusions for Experiment-3: 1) For offline applications, which only care about quality and neglect time consumption, \textit{Del} is the best choice, and our \textit{ImMesh} is the second best one. 2) For real-time applications, our work \textit{ImMesh} is the best choice. Even though \textit{TSDF} with GPU acceleration can run in real-time, its meshing correctness is much lower than \textit{ImMesh}.  


\begin{figure*}[t]
    \centering
    \setlength{\belowcaptionskip}{-0.5cm}
    \includegraphics[width=1.0\linewidth]{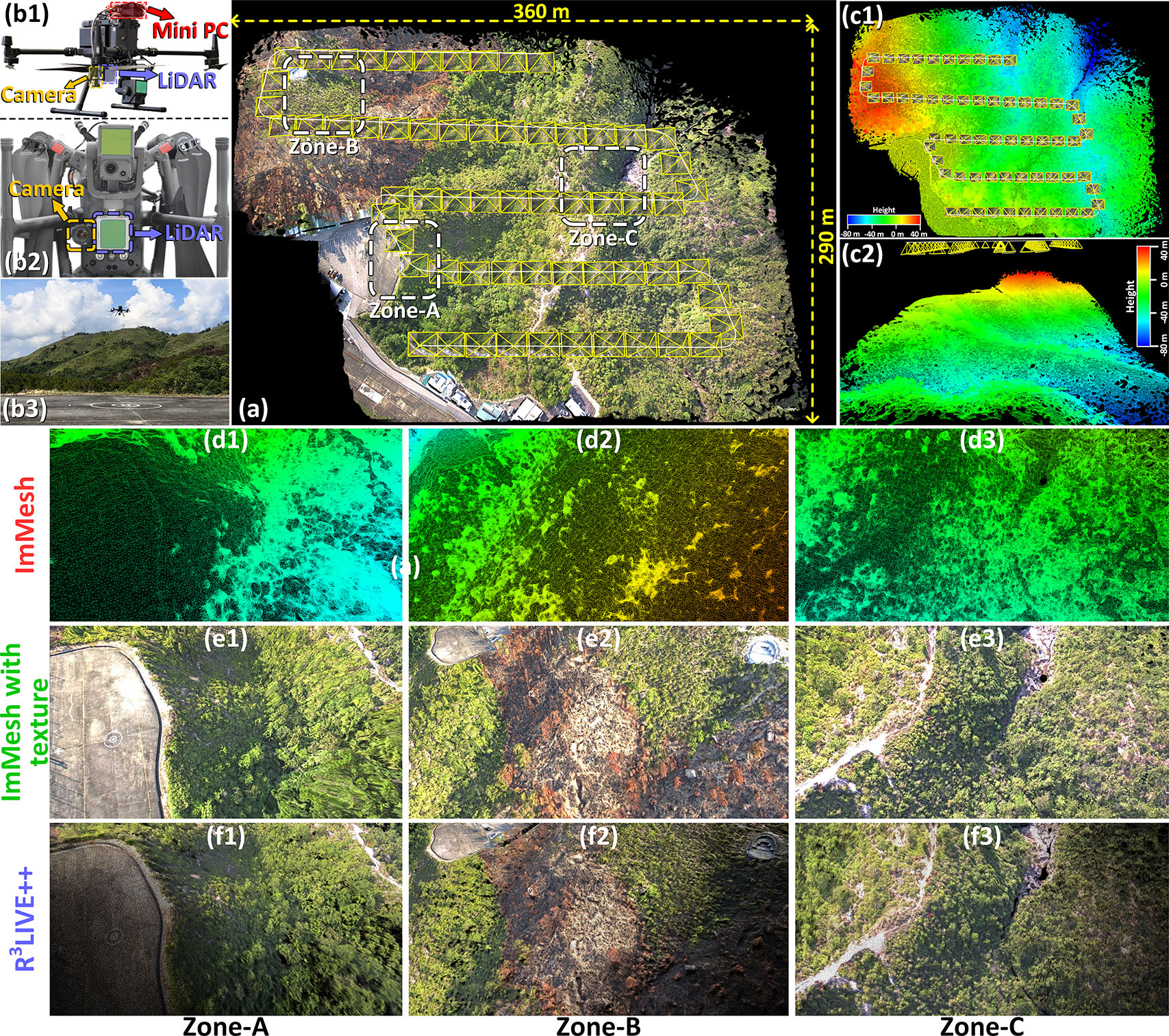}
    \caption{(b1$\sim$b3) show our UAV platform for data collection. (a) show the bird view of our lossless texture reconstruction result. (c1 and c2) show the altitude of this map by coloring the facets in their height w.r.t. the take-off point (i.e., the ground plane in Zone-A). The qualitative comparison of mapping results in Zone-A, B, and C of ImMesh, ImMesh after texturing, and R$^3$LIVE++ are shown in (d$\sim$f). To see the detailed reconstruction process of the scene, please refer to our \href{https://youtu.be/pzT2fMwz428?t=622}{\VIDEO{accompanying video \cite{immeshvideo} (starting at 10:22)}} on YouTube.} 
    \label{fig_app_fast_surveying}
     \vspace{0.3cm}
\end{figure*}





\subsection{Application-1: LiDAR point cloud reinforcement}\label{Application_1_LiDAR_reinforce}
Benefiting from ImMesh's real-time ability to reconstruct the triangle mesh on the fly, depth images can be rasterized from the reconstructed facets online in the current sensor frame. By unprojecting the 3D points from the depth image, point clouds of a regular pattern can be retrieved with wider FoV and denser distribution than the original input LiDAR scan. We termed this process as LiDAR point reinforcement.

In this experiment, we demonstrate the LiDAR point cloud reinforcement with a solid-state LiDAR \textit{Livox Avia} with FoV of $70.4^\circ \times 77.2^\circ$. The comparisons between the original points of a LiDAR frame (colored in white) and after our reinforcement (colored in magenta) with different sets of rasterization FoV are shown in Fig. \ref{fig_LiDAR_point_cloud_reinforcement}. As the white points shown in the first row of Fig. \ref{fig_LiDAR_point_cloud_reinforcement}, the input LiDAR scan is sparse with an irregular scanning pattern. After the reinforcement, the resultant 3D points colored in magenta are distributed in a regular pattern, with a higher density and wider FoV (as the rasterization FoV is bigger than LiDAR's). To better understand their differences, we present the comparisons of depth images after projection, as shown in the second and third rows of Fig. \ref{fig_LiDAR_point_cloud_reinforcement}. 


\subsection{Application-2: Rapid, lossless texture reconstruction}\label{sect_application_2}
In this application, we show how ImMesh can be applied in applications of lossless texture reconstruction for rapid field surveying. As shown in Fig. \ref{fig_app_fast_surveying}(b1$\sim$b3), we mounted a \textit{Livox avia} LiDAR and a \textit{Hikvision CA-050-11UC} global shutter RGB camera on a \textit{DJI M300} drone platform.

We collected the data in a mountain field by taking off from Zone-A (see Fig. \ref{fig_app_fast_surveying}(a)) and flying in a ``s"-like pattern trajectory with a traveling distance of \SI{975}{\meter}. We leveraged ImMesh for reconstructing the mesh from the collected LiDAR data and used R$^3$LIVE++ \cite{r3livepp} for estimating the camera's poses (as the yellow frustum shown in Fig. \ref{fig_app_fast_surveying}(a, c1 and c2)). We textured each facet of the reconstructed mesh by the RGB image captured by the nearest camera frame with the estimated camera pose from R$^3$LIVE++. Benefiting from the high efficiency of ImMesh and R$^3$LIVE++, the total time of reconstructing the RGB textured mesh from this sequence of duration \SI{325}{\second} cost only \SI{686}{\second}, with \SI{328}{\second} for ImMesh, \SI{330}{\second} for R$^3$LIVE++, and \SI{28}{\second} for texturing. Fig. \ref{fig_app_fast_surveying}(a) shows a bird view of our mesh after texturing, with the close-up views of textured mesh in Zone-A, B, and C shown in Fig. \ref{fig_app_fast_surveying}(e1, e2, and e3), respectively. In Fig. \ref{fig_app_fast_surveying}(c1 and c2), we show the altitude of this map by coloring the facets in their height w.r.t. the take-off point (i.e., the ground plane in Zone-A). 

As shown by the close-up views in the bottom three rows of Fig. \ref{fig_app_fast_surveying}, the reconstructed mesh (d1$\sim$d3) from our ImMesh after texturing (e1$\sim$e3) successfully preserves the map textures when comparing with the RGB-colored point cloud reconstructed by R$^3$LIVE++ (f1$\sim$f3). Due to the limited point cloud density, the RGB-colored point cloud by R$^3$LIVE++ is unable to reconstruct the scene losslessly. Compared to existing counterparts (e.g., 3D reconstruction from photogrammetry\cite{litvinov2013incremental, schonberger2016structure}) that reconstructs a scene from captured images (and RTK measurements), our system shows significant advantages: 1) It is a reliable solution that does not require GPS measurement. 2) It is a rapid reconstruction method that costs only 2$\sim$3 times the data sampling time for reconstructing a scene. 3) It preserves a geometry structure of high accuracy that is reconstructed from LiDAR's measurements.
 The \href{https://youtu.be/pzT2fMwz428?t=622}{\VIDEO{accompanying video \cite{immeshvideo} (starting at 10:22)}} that records the full process of this lossless texture reconstruction is available on our YouTube, and an additional trial is shown in our Supplementary Material\cite{immeshsup}.

Notice that in Fig. \ref{fig_app_fast_surveying},  the presence of isolated mesh facets is a result of missing scanning data, while the blurry texture artifacts are caused by the large viewing angle of the facets and textured images, both can be addressed through proper data collection processes. 

\section{Conclusions and future work}
\subsection{Conclusions}
In this work, we proposed a novel meshing framework termed ImMesh for achieving the goal of simultaneous localization and meshing in real-time. The real-time incremental meshing nature of our system, even in large-scale scenes, makes it one of a kind. The \textit{localization} module in ImMesh represents the surrounding environment in a probabilistic representation, estimating the sensor pose in real-time by leveraging an iterated Kalman filter to maximize the posterior probability. The \textit{meshing} module directly utilizes the spatially-downsampled registered LiDAR points as mesh vertices and reconstructs the triangle facets in a novel incremental manner in real-time. To be detailed, our \textit{meshing} module first retrieves all voxels that contain newly appended vertices. Then, the voxel-wise 3D meshing problem is converted into a 2D one by performing dimension reduction for efficient meshing. Finally, the triangle facets are incrementally reconstructed with \textit{pull}, \textit{commit}, and \textit{push} steps.

Our system is evaluated by real experiments. First, we verified the overall performance by presenting live video demonstrations of how the mesh is immediately reconstructed in the process of data collection. Then we extensively tested ImMesh with four public datasets collected by four different LiDAR sensors in various scenes, which confirmed the real-time ability of our system. Lastly, we benchmarked the meshing performance of ImMesh in Experiment-3 by comparing it against existing meshing baselines. The results show that ImMesh achieves high meshing accuracy while keeping the best runtime performance among all methods. 

Applications of our system were demonstrated. We first show how ImMesh can be applied for LiDAR point cloud reinforcement, which generates reinforced points in a regular pattern with denser density and wider FoV than raw LiDAR scans. In Application-2, we combined our works ImMesh and R$^3$LIVE++ to achieve the goal of lossless texture reconstruction of scenes. Finally, we make our code publicly available on our GitHub: \href{https://github.com/hku-mars/ImMesh}{\tt github.com/hku-mars/ImMesh}.

\subsection{Limitations and future works}

One major limitation of our work is its lack of scalability in spatial resolution. Specifically, when dealing with large planar surfaces, ImMesh tends to inefficiently reconstruct the mesh with numerous small facets due to the fixed vertex density. Conversely, for tiny objects smaller than the size of a voxel, ImMesh struggles to accurately reconstruct their surfaces, as mentioned in our quantitative evaluation results in Section \ref{sect_exp3_res_ana_accuracy}. To address this limitation, our future work will focus on developing an adaptive resolution meshing strategy.

The second limitation is that our system does not currently implement any loop correction mechanism, resulting in potential gradual drift due to accumulated localization errors at revisited places. This potentially leads to inconsistent reconstructed results if revisit occurs. In our future work, we plan to address this limitation by integrating our recent works \cite{yuan2022std, lin2019fast} on loop detection based on LiDAR point clouds. This loop detection mechanism will allow us to detect loops online and apply loop corrections to reduce drift and improve the consistency of the reconstructed results.


Furthermore, we have noticed that a number of works appeared in the literature recently, which utilize the reconstructed mesh for improving the localization accuracy of both visual-slam (e.g., \cite{panek2022meshloc}) and LiDAR-slam system (e.g., \cite{vizzo2021poisson, dreher2021global, oelsch2021r}). Motivated by these works, our future work would improve our localization accuracy by utilizing our online reconstructed mesh.

Lastly, when realizing the goal of lossless texture reconstruction of scenes, we combined ImMesh and R$^3$LIVE at the system level as presented in our Application-2 (in Section \ref{sect_application_2}). Our future would couple ImMesh with R$^3$LIVE more tightly to improve the overall efficiency. 

\section{Acknowledgements}
The authors would like to thank DJI Co., Ltd\footnote{\url{https://www.dji.com}} for providing devices and research funds.

\fi

\bibliography{ImMesh}


\clearpage
\setcounter{equation}{0}
\setcounter{figure}{0}
\setcounter{table}{0}
\setcounter{page}{1}
\setcounter{section}{1}
\setcounter{section}{0}%
\setcounter{subsection}{0}%
\setcounter{subsubsection}{0}%
\setcounter{paragraph}{0}%
\onecolumn
\begin{landscape}
\begin{figure}[ht]
\begin{centering}
    \begin{minipage}{0.99\linewidth}
        \nopagebreak
        \centering
        \caption*{\textbf{\Large Supplementary Material: An additional trial of our lossless texture reconstruction based on ImMesh}}
        \vspace{1.2cm}
        \includegraphics[width=1.0\linewidth]{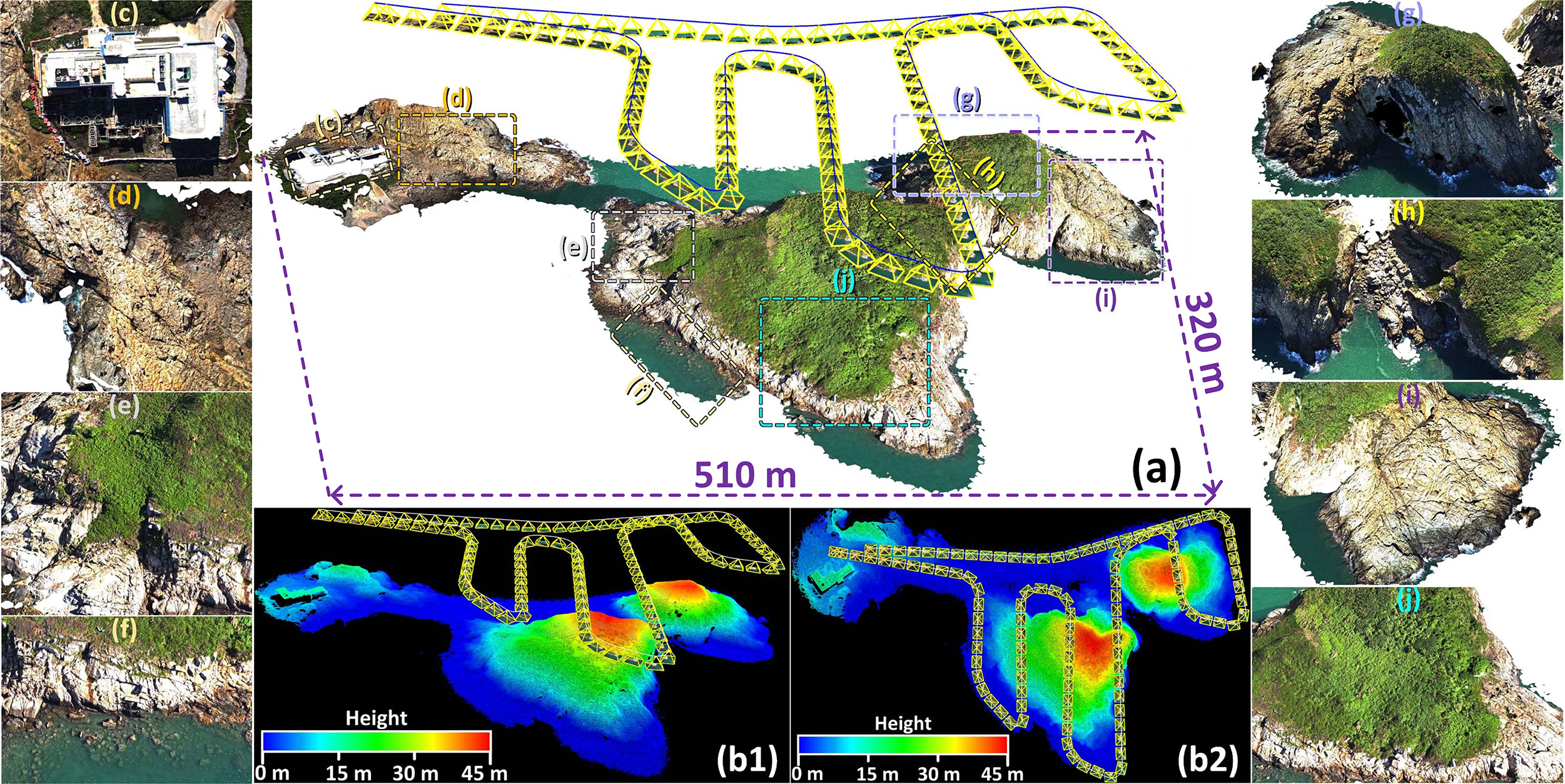}
        \vspace{-0.3cm}
        \caption{Results of an additional trial test. In this trial, we collected the data by flying over islands in an ``B"-like trajectory, as shown by the blue path in (a). (b1) and (b2) show the side view and bird view of our reconstructed triangle mesh, where the mesh is colored by their altitude w.r.t. the sea level. (a) show the overview of our lossless texture reconstruction result, where we use the estimated camera poses (the yellow frustums) of R$^3$LIVE++ for texturing the mesh with the collected images. The entire texture reconstruction of this \SI{578}{\second} sequence only costs \SI{1210}{\second} (on \textit{Intel i9-10900}), with \SI{583}{\second} for ImMesh, \SI{587}{\second} for R$^3$LIVE++, and \SI{40}{\second} for texturing. To see the detailed reconstruction process of the scene, please refer to our video on YouTube: \href{https://youtu.be/pzT2fMwz428?t=892}{\tt{youtu.be/pzT2fMwz428?t=892}}.}
        \label{fig:bear}
    \end{minipage}
\end{centering}
\end{figure}
\end{landscape}
\end{document}